\documentclass[twoside]{ctuthesis}

\ctusetup{
	mainlanguage = english,
	otherlanguages = {slovak,english},
	title-czech = {Vylepšení HardNet deskriptoru},
	title-english = {Improving the HardNet Descriptor},
	doctype = M,
	faculty = F3,
	department-czech = {Katedra kybernetiky},
	department-english = {Department of Cybernetics},
	author = {Bc. Milan Pultar},
	supervisor = {M.Sc. Dmytro Mishkin},
	supervisor-address = {Czech Technical University in Prague, \\ Faculty of Electrical Engineering, \\ Department of Cybernetics, \\ Karlovo namesti 13, \\ 121 35 Prague 2, \\ Czech Republic},
	fieldofstudy-english = {Open Informatics},
	subfieldofstudy-english = {Computer Vision and Image Processing},
	fieldofstudy-czech = {Otevřená informatika},
	subfieldofstudy-czech = {Počítačové vidění a digitální obraz},
	keywords-czech = {deskriptor bodu zájmu, HardNet, registrace obrázků, konvoluční neuronová síť, robustnost ke změně osvětlení, tvorba datasetu, redukce datasetu, kombinování datasetů, hledání architektury, ztrátová funkce, komprese vektorové reprezentace},
	keywords-english = {local feature descriptor, HardNet, image matching, convolutional neural network, illumination robust, dataset creation, dataset reduction, combining datasets, architecture search, loss function, compression of embeddings},
	day = 22,
	month = 5,
	year = 2020,
	specification-file = {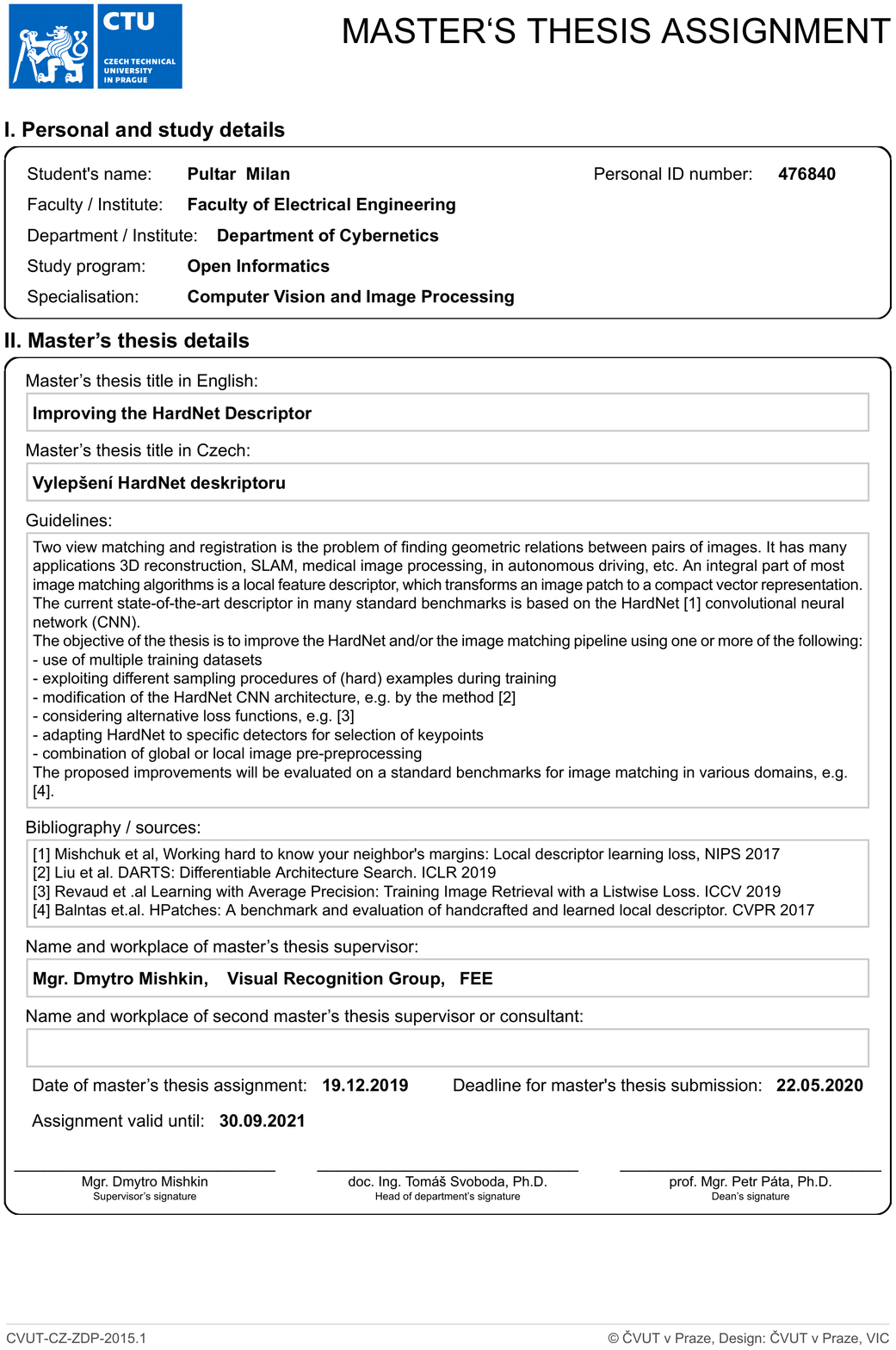},
	front-specification = true,
}

\ctuprocess

\addto\ctucaptionsczech{%
}

\ctutemplateset{maketitle twocolumn default}{
	\begin{twocolumnfrontmatterpage}
		\ctutemplate{twocolumn.thanks}
		\ctutemplate{twocolumn.declaration}
		\ctutemplate{twocolumn.abstract.in.titlelanguage}
		\ctutemplate{twocolumn.abstract.in.secondlanguage}
		\ctutemplate{twocolumn.tableofcontents}
		\ctutemplate{twocolumn.listoffigures}
	\end{twocolumnfrontmatterpage}
}

\usepackage[pagewise]{lineno}
\usepackage{lipsum,blindtext}
\usepackage{times}
\usepackage{epsfig}
\usepackage{graphicx}
\usepackage{amsmath}
\usepackage{amssymb}
\usepackage{booktabs}
\usepackage{enumitem}
\usepackage{float}
\usepackage{capt-of}
\usepackage{tabu}
\usepackage{tabularx}
\usepackage{booktabs}    % professional-quality tables
\usepackage[ampersand]{easylist}
\usepackage{subcaption}
\usepackage[export]{adjustbox}
\usepackage{siunitx}
\usepackage[bottom]{footmisc}
\usepackage{mathrsfs} % provides \mathscr used in the ridiculous examples
\usepackage{verbatim}
\usepackage{bbding}
\usepackage{pifont}
\usepackage{wasysym}
\usepackage{amssymb}
\usepackage{gensymb}

\theoremstyle{plain}

\theoremstyle{definition}

\theoremstyle{note}
\newtheorem*{remark*}{Remark}

\begin{abstract-czech}
Uvažujeme problém učení lokálního deskriptoru pro wide baseline stereo, zaměřeujeme se na deskriptor HardNet, který je blízko state-of-the-art. Představujeme datový soubor AMOS Patches, který zlepšuje odolnost vůči změnám osvětlení a vzhledu. Je založen na registrovaných obrázcích z vybraných kamer z datového souboru AMOS. Dáváme doporučení k procesu vytváření datových sad obrázkových výřezů a testujeme HardNet naučený na datech různého charakteru. Představujeme také metody pro kombinaci datových sad a jejich redukci, díky kterým dává učení na výrazně méně datech srovnatelné výsledky.

HardNet8, který překonává původní HardNet, těží z provedených rozhodnutí o architektuře: schéma propojení, final polling, receptivní pole, stavební bloky CNN nalezené pomocí manuálních nebo automatických vyhledávacích algoritmů - DARTS. Ukazujeme vliv často přehlížených hyperparametrů, jako je velikost batche a délka tréninku, na kvalitu deskriptoru. Komprese výstupů sítě pomocí PCA dále zlepšuje výsledky a také snižuje paměťovou náročnost.

Na základě získaných poznatků představujeme dvě varianty HardNet8 deskriptoru: jeden má dobré výsledky na HPatches, AMOS Patches a IMW Phototourism, druhý je optimalizován pro IMW Phototourism.

\par\ \par\ \par\ \par\ \par\ \par\ \par\ \par\ \par\ 

\end{abstract-czech}

\begin{abstract-english}
We consider the problem of local feature descriptor learning for wide baseline stereo focusing on the HardNet descriptor, which is close to state-of-the-art. AMOS Patches dataset is introduced, which improves robustness to illumination and appearance changes. It is based on registered images from selected cameras from the AMOS dataset. We provide recommendations on the patch dataset creation process and evaluate HardNet trained on data of different modalities. We also introduce a dataset combination and reduction methods, that allow comparable performance on a significantly smaller dataset.

HardNet8, consistently outperforming the original HardNet, benefits from the architectural choices made: connectivity pattern, final pooling, receptive field, CNN building blocks found by manual or automatic search algorithms -- DARTS. We show impact of overlooked hyperparameters such as batch size and length of training on the descriptor quality. PCA dimensionality reduction further boosts performance and also reduces memory footprint. 

Finally, the insights gained lead to two HardNet8 descriptors: one performing well on a variety of benchmarks -- HPatches, AMOS Patches and IMW Phototourism, the other is optimized for IMW Phototourism.

\par\ \par\ \par\ \par\ \par\ \par\ \par\ \par\ \par\ \par\ 

\end{abstract-english}

\begin{thanks} % Acknowledgements / Podekovani
I wish to express my sincere appreciation to my supervisor Dmytro Mishkin for his support and patient guidance. He educated me in the fields of computer vision and helped me pursue my goals. I would also like to thank Jiří Matas for his enthusiastic encouragement and for giving me valuable advice in academia as well as in day-to-day life. I would like to extend my thanks to all members of CMP for creating a friendly and inspiring environment. Finally, I would like to thank Nicole for her love and kind support during my studies.
\end{thanks}

\begin{declaration} % Declaration / Prohlaseni
I declare that the presented work was de-
veloped independently and that I have
listed all sources of information used
within it in accordance with the methodical instructions for observing the ethical
principles in the preparation of university
theses.

Prague\hfill 22 May 2020

\strut \hfill ............................... \newline
\strut \hfill signature

\end{declaration}

\begin{document}
\maketitle

\chapter{Introduction}
\label{sec:introduction}
We consider the wide baseline stereo, where the goal is to find correspondences between pixels and regions and estimate the relative camera pose from one image to the other. The list of applications is vast, it can be applied in the tasks of structure from motion (SfM)~\cite{schonberger2016structure, torii2018structure}, image retrieval~\cite{shen2018matchable}, relocalization~\cite{zhou2019learn, sarlin2019coarse}, simultaneous localization and mapping, self-localization and in applications like autonomous driving~\cite{dewan2018learning} and panorama stitching~\cite{brown2007automatic}. The pipeline consists of several steps. First, feature points are detected in the two input images, then small image patches are cropped around these points. Patches are then normalized with respect to rotation, scale and possibly other factors like contrast. A descriptor maps to a vector representation for each of the patches. Subsequently tentative correspondences are created between the two sets of descriptors via nearest neighbor matching in the descriptor space and verified by RANSAC~\cite{consensus1981paradigm} and the geometry relation is then obtained from the set of inliers~\cite{hartley1994projective}. Two popular choices of an image relation model are homography -- for the planar surfaces -- and epipolar geometry, which are represented by essential or fundamental matrices \cite{Hartley2004}.

\begin{figure}[htb]
\centering

\includegraphics[width=0.99\linewidth]{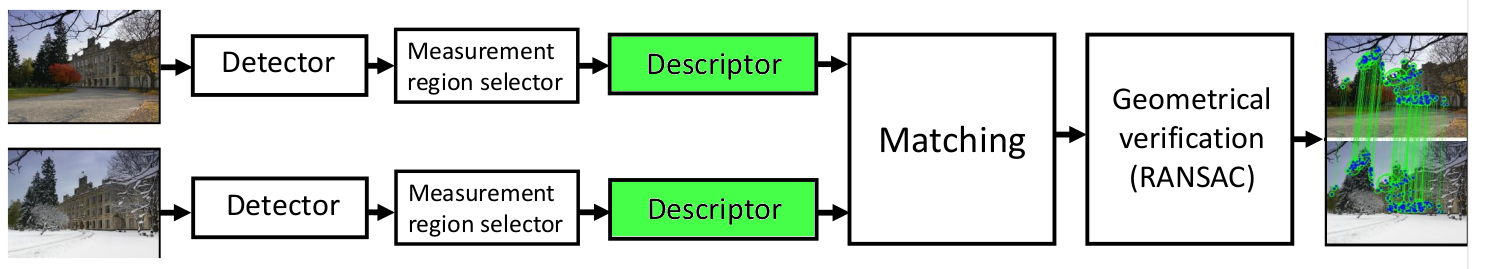}

\caption[The image matching pipeline.]{The wide baseline stereo pipeline. The stage, which is studied in current work is highlighted. Image credit: CTU Computer Vision Methods course \footnotemark .}
\label{fig:wbs}
\end{figure}

A local feature descriptor is a mapping from a (small) image to a metric space, which is robust to changes in acquisition conditions so that descriptors related to the same 3D points are similar and dissimilar otherwise. A descriptor is typically used to establish tentative correspondences between detected local features like blobs, corners or more complex structures. Performance of the whole WBS pipeline relies heavily on the qualities of the descriptor, namely its performance, speed, universality and robustness to changes in acquisition conditions.

\footnotetext{https://cw.fel.cvut.cz/wiki/courses/mpv/start}

The earliest descriptors were handcrafted. One of the most famous is SIFT \cite{lowe2004distinctive}, which computes a histogram of weighted gradient orientations to form a 128-dimensional descriptor. Other descriptors~\cite{LIOP2016,MROGH2011} compute the relative ordering of pixel values. Recent years have witnessed a noticeable effort to move to descriptors obtained by deep learning~\cite{mishchuk2017working,l2net}. Existing work explore possible architectures~\cite{l2net}, loss functions~\cite{tfeat2016,mishchuk2017working,He2018CVPR,Keller2018CVPR} and improvements of robustness to viewpoint changes by introducing large scale datasets from 3D reconstruction~\cite{mitra2018large, geodesc2018}.

In some works there has been an effort to train detector and descriptor as a single model~\cite{noh2017large,revaud2019r2d2,dusmanu2019d2}. Although these methods were able to perform well in the most recent benchmark~\cite{Jin2020}, they are still worse than the combination of handcrafted DoG detector (used in SIFT) and HardNet-like descriptor. We focus on improving the descriptor part, namely using the HardNet architecture~\cite{mishchuk2017working} with the triplet margin loss function.

Regarding acquisition conditions, we refer to~\cite{Mishkin2015WXBS} for an introduction to wide-baseline matching and an overview of 4 main factors - sensor type, geometry, illumination, and appearance - which influence the performance of the descriptor. In this work we focus on the last three, that is we want to improve robustness to changes in the intensity, location and direction of light sources; to change in camera pose, its scale and resolution; to season-related changes and changes in visibility in the scene (fog, snow).

Robustness to illumination and appearance changes has received comparatively little attention, yet it is a bigger challenge for modern descriptors~\cite{wxbs2015,balntas2017hpatches}. Their performance degrades significantly if the images are taken at different times of day, different seasons or weather conditions \cite{Verdie2015}. We tackle this problem by leveraging information from 24/7 webcams located worldwide~\cite{jacobs2007consistent,jacobs2009global}.

\begin{figure}[htb]
\centering

\includegraphics[max width=1\textwidth]{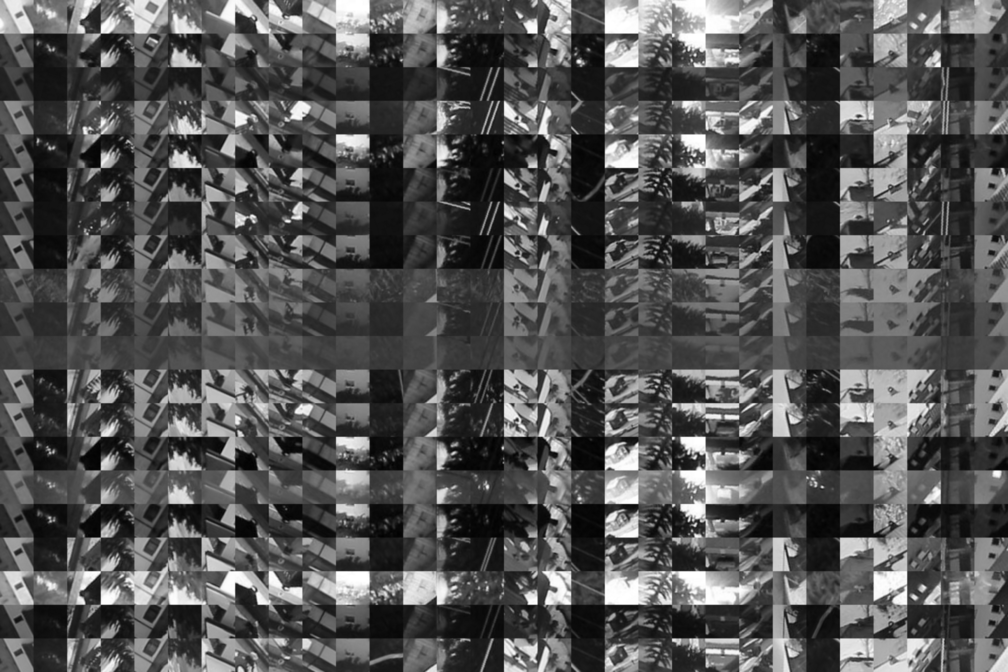}

\caption[A subset of the AMOS Patches dataset.]{A subset of the AMOS Patches dataset originating from 24/7 webcams located world-wide. Each column contains 20 corresponding patches.}
\label{fig:patches}
\end{figure}

We make the following contributions. In Section \ref{sec:creating_patches} we present a method for selecting suitable static views for the training of a descriptor together with a fixed number of the most representative images from a (too) large collection of camera feeds. We also introduce AMOS Patches\footnote{The dataset and contributing images are available at \url{https://github.com/pultarmi/AMOS_patches}} - extracted image crop-outs forming sets of veridical patch correspondences, see Figure \ref{fig:patches}. This dataset improves robustness to changes in illumination and appearance of the trained descriptor.

In the second part of Section \ref{sec:extracting} we focus on further improving the HardNet \cite{mishchuk2017working} descriptor by making adjustments to data preparation and training process. In Sections \ref{sec:expsamos} and \ref{sec:expsimw} we present experiments with AMOS Patches and IMW Phototourism~\cite{Jin2020}. They are related to data modality and the number of source views. We also introduce a method for dataset reduction and compare a few ways of combining datasets.

% Numerous experiments were carried out to assess the benefits of several data modalities. 

In Section \ref{sec:architectures} we conduct a systematic study on the influence of different architectural and choices on the descriptor performance: final pooling, architecture style, receptive field size.

In Section \ref{sec:compression} we explain our approach to compressing network outputs that increases the descriptor performance.

Section \ref{sec:loss} presents experiments and choices related to the loss function. Hyperparameters are set in section \ref{sec:hyper}. We summarize our findings in Section \ref{sec:conclusion}.

Our implementation uses PyTorch~\cite{paszke2017automatic} package and fastai library~\cite{howard2020fastai}.

Part of the following work is based on the published paper \cite{HardNetAMOS2019}. The main author is also author of this thesis.

\chapter{Related work}
\label{chap:related_work}

\begin{comment}
Structure:
1) earliest descriptors, handcrafted, what are their advantages in general
1a) binary descs
2) deep learned
3) how can we improv edeep learned
- architecture, loss, sampling, dataset
4) datasets
 - train, test
 
mention illumination invariance
mention: PCA, BRISK? 
\end{comment}

The literature on local feature descriptors is vast \cite{Csurka2018}. The earliest descriptors were handcrafted. One of the most widely used is SIFT~\cite{lowe2004distinctive}, which uses a weighted histogram of gradient orientations and is moderately robust to local deformations and intensity changes\cite{Mikolajczyk2005}. DSP-SIFT~\cite{dong2015domain} is a modification, which uses also pooling of gradient orientations across different scales. 
% HalfSIFT~\cite{HalfSIFT2007} does not differentiate opposite directions, introducing invariance to contrast reversal.

The family of order-based descriptors like LIOP~\cite{LIOP2016} or MROGH~\cite{MROGH2011} operates on the relative order of pixel intensities in the patch instead of on the intensities themselves. Relative order (sorting) is invariant to any monotonically increasing intensity transformation. Descriptors like SymFeat~\cite{Hauagge2012}, SSIM~\cite{Shechtman2007} and learned DASC~\cite{DASC2017} encode local symmetries and self-similarities.

There are two main ways towards achieving robustness to illumination change: by descriptor construction and by learning on the appropriate dataset.  Normalization of the patch mean and variance is a simple but powerful method, which is implemented in both SIFT~\cite{lowe2004distinctive} and modern learned descriptors~\cite{l2net, mishchuk2017working}. It makes the descriptor invariant to affine changes in pixel intensities in the patch. Photometric normalization might be more complex than mean-var normalization, e.g.  as done by the learned RGB2NIR~\cite{Zhi_2018_CVPR} or hand-crafted LAT~\cite{LAT2017}. HalfSIFT~\cite{HalfSIFT2007} treats opposite intensity gradient directions as equal, trading off half of the SIFT dimensionality for being contrast reversal invariant. It works well in medical imaging and infrared-vs-visible matching.

Binary descriptors were developed to fill the demand for compact and fast representations, although they have usually worse performance than the real-valued ones. BRIEF~\cite{calonder2010brief} and BRISK~\cite{leutenegger2011brisk} construct a descriptor based on comparison of pairs of randomly selected pixels. ORB~\cite{rublee2011orb} is an improvement of BRIEF with an estimation of patch orientation using FAST local feature point detector. The similarity between descriptors is computed efficiently with Hamming distance. L2 distance is commonly used in the case of real-valued descriptors.

Current research on local feature descriptors is shifting towards those obtained by deep learning. Significant advancement has been made in recent years~\cite{balntas2017hpatches,Jin2020}. In the following text we focus on real-valued ones. A simple three-layer network was used in~\cite{simo2015discriminative} together with a stochastic sampling of hard examples in a metric learning approach. L2Net~\cite{l2net} introduced relatively simple convolutional network architecture, which outputs a vector of length 128. HardNet \cite{mishchuk2017working} adopted the L2Net architecture and introduced the triplet loss function with hard-in-batch sampling. GeoDesc~\cite{geodesc2018} uses geometry constraints from multi-view reconstructions. SOSnet~\cite{tian2019sosnet} introduced second order similarity regularization, which is added to the loss function.

Learned descriptors for multimodal data, e.g. infrared-vs-visible mostly include Siamese convolution networks with modality-specific branches, like the Quadruplet Network~\cite{Qnet2017}. The decision which branch to use for a specific patch comes from an external source or a domain classifier. HNet~\cite{HNet2018} uses an auto-encoder network and style transfer methods like CycleGAN~\cite{CycleGAN2017} for emulating different modalities.

There are many datasets potentially suitable for training and testing of local feature descriptors. Most of them comprise images only, i.e. without patch-level correspondences. DTU Robot~\cite{dturobot2012} contains photos of objects in a lab environment, camera positioning is operated by a robotic arm, light sources are artificial. RobotCar dataset~\cite{RobotCarDatasetIJRR} contains camera feed from a vehicle, which was repeatedly traversing a predefined route in central Oxford. TILDE~\cite{Verdie2015} dataset comprises images from 6 scenes, it was made for training and evaluation of illumination-robust detector. Other datasets are OxfordAffine~\cite{Mikolajczyk2005}, Aachen Day-Night~\cite{Sattler2018}, GDB~\cite{yang2007registration}, SymBench~\cite{Hauagge2012}, etc. However, none of these datasets contain also keypoint correspondences, which are needed for the extraction of patches.

%In the current section we will describe existing datasets and the one we present.

Despite the importance of the topic, the number of patch-level datasets is small, especially for training and testing illumination-robust descriptors. UBC Phototour dataset~\cite{brown2007automatic} contains patches from 3 scenes, it covers more viewpoint related changes. HPatches dataset ~\cite{balntas2017hpatches} contains 116 scenes, each represented by 6 images. It is the only patch-based dataset we know of which contains sequences with illumination changes. 57 scenes contain 3D objects under varying photometric conditions, while 59 scenes are planar with changes in camera pose. The PhotoSynth (PS) dataset~\cite{mitra2018large} comprises 30 scenes reconstructed by the COLMAP~\cite{schonberger2016structure} SfM software. Although the scenes contain photometric changes, the trained descriptor achieves better results only on the viewpoint split of HPatches.

\chapter{HardNet Descriptor}

\begin{figure}[htb]
\centering

\includegraphics[width=0.99\linewidth]{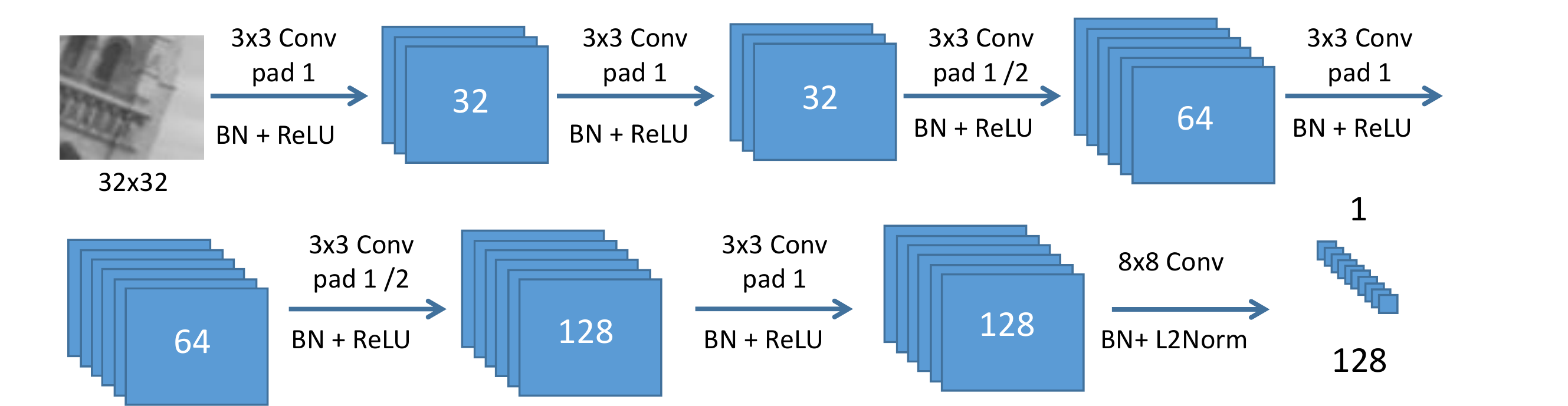}

\caption[HardNet descriptor architecture.]{HardNet descriptor architecture.  Image credit:~\cite{mishchuk2017working}.}
\label{fig:hardnetarch}
\end{figure}

We have selected the HardNet descriptor~\cite{mishchuk2017working} as an object to study and improve given its good performance in various benchmarks~\cite{Jin2020}.
HardNet descriptor is a real-valued convolutional network with the output vector of length 128 and the input patch of size 32$\times$32p pixels. The architecture adopts the VGG~\cite{simonyan2014very} style, it is composed of convolutional layers, each followed by batch normalization and ReLU activation. Dropout with $p=0.3$ is used. The inputs to the network are mean and variance normalized.

% The first descriptors were handcrafted. Currently the most widely used is SIFT~\cite{lowe2004distinctive}, which computes 16 local histograms of gradient orientations in a 16x16 grid of pixels and concatenates them to a 128-dimensional descriptor. It has been empirically found to be robust to illumination changes, small shifts and affine changes. There are also several variants that increase robustness to changes in illumination~~\cite{HalfSIFT2007} or scale~\cite{dong2015domain}. 

The task of learning a local feature descriptor is formulated as a metric learning problem, where the goal is to learn an embedding for the input patches such that corresponding (positive) patches are closer than the noncorresponding (negative) ones. For training we use the hard-in-batch triplet margin loss~\cite{mishchuk2017working}. Let $(p_{i,1}, p_{i,2}), i=1,\dots,n$ be pairs of corresponding patches in a batch of size n. Let $d$ be the Euclidean norm. The loss function is the following:

% The task of learning a local feature descriptor can be formulated as a metric learning problem, where the goal is to learn an embedding for the input patches such that corresponding (positive) patches are closer than the noncorresponding (negative) ones. For training we use the hard-in-batch triplet margin loss~\cite{mishchuk2017working}. Let $\{a_i\}_{i=1}^n, \{p_i\}_{i=1}^n$ be two sequences of patches in a batch of size n, where a patch in the position $i$ in the first sequence is from the same patch set as the patch in the same position in the second sequence. $\{a_i\}_{i=1}^n$ are called anchors, $\{p_i\}_{i=1}^n$ are positives. Let $d$ be the Euclidean norm. The loss is then computed as

\begin{align}
L &= \frac{1}{n}\sum\limits_{i=1,n} \max(0,M + d_{\textrm{pos}}^i - \min(d_{\textrm{neg1}}^i, d_{\textrm{neg2}}^i)), \label{eq:lossfc}\\
d_{\textrm{pos}}^i &=d(p_{i,1},p_{i,2}), \nonumber \\
d_{\textrm{neg1}}^i&=\min_{j=1,..n, j\neq i} d(p_{i,1}, p_{j,2}), \nonumber \\
d_{\textrm{neg2}}^i&=\min_{j=1,..n, j\neq i} d(p_{j,1}, p_{i,2}), \nonumber
\end{align}

where M=1 is called the margin. By minimizing the loss we aim to minimize the distance between the positives while maximizing the distance to the hardest negatives from both positive patches. The loss is zero if the difference between the latter and the former distance is at least M. 

\section{Local Descriptor Evaluation Methods}

\begin{figure*}[htb]
\centering

\includegraphics[width=0.9\linewidth]{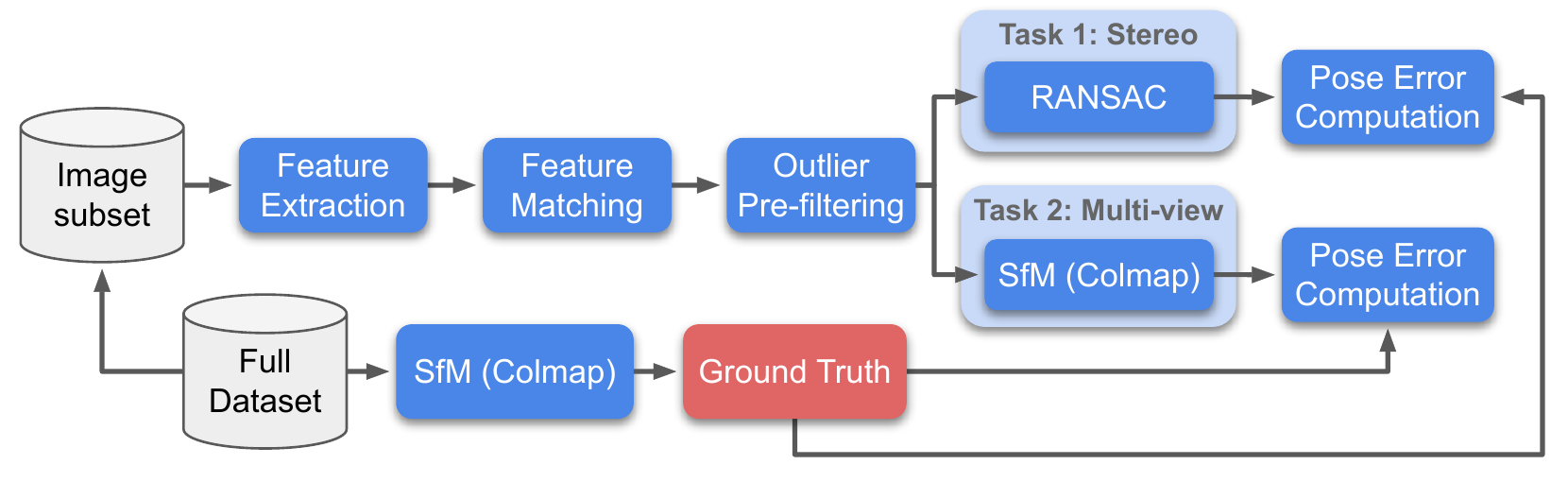}

\caption[Pipeline of the IMW PT benchmark.]{Pipeline of the IMW PT benchmark. The input is a set of images. It detects keypoints, extracts features and computes matches. Camera pose is then computed after running outlier filtering and Random sample consensus (RANSAC). Image credit:~\cite{Jin2020}.}
\label{fig:cvprbench}
\end{figure*}

There are two types of evaluation. One way is to evaluate either solely the descriptor using pre-extracted patches, the second way is to evaluate some end-goal-metric of the whole image matching pipeline by plugging in the descriptor and computing the precision of the pipeline as a whole. An example of end-goal-metric is camera pose accuracy. We describe and use both approaches.

The important property of the first approach is zero dependence on other parts of the image matching pipeline. The procedure is generally simpler, faster and we can make clear judgments on the descriptor. On the other hand, it provides us limited information on how the image matching pipeline will perform with the descriptor plugged in. One of the used metrics is a false positive rate (FPR), which measures the percentage of false positives at a given recall level. Specifically, FPR95 - FPR at 0.95 recall -- is usually used in UBC Phototour. It is defined as

\begin{align}
    \textrm{FPR95} = \frac{\textrm{FP}}{\textrm{FP}+\textrm{TN}},\ \textrm{s.t.}\  \frac{\textrm{TP}}{\textrm{TP}+\textrm{FN}}=0.95,
\end{align}
considering ascending ordering of samples according to assigned confidence score.
We observed this metric has little correlation with the actual performance of the descriptor and so its usability is limited to a sanity check of convergence during the training of a model.

% This metric has low informative value and we observed that it has little correlation with the actual performance of the descriptor. It can be used as a sanity check of converge during the training of a model.

Another metric is mean average precision (mAP), formulated as 
\begin{align}
    \textrm{mAP} &= \frac{1}{Q} \sum\limits_{i=1}^Q \textrm{AP}_i,\\
    \textrm{AP}_i &= \sum\limits_{j=1}^{N_i} P(s_{i,j}) \nabla r(s_{i,j}), \nonumber
\end{align}
where $s_{i,j}$ is the $j$-th largest assigned confidence score in query $i$, $\nabla$ denotes the difference in recall between samples $s_{i,j}$ and $s_{i,j-1}$, $P$ denotes precision, considering samples $s_{i,1},\dots,s_{i,j}$. AP$_i$ values are averaged over $Q$ queries, where each query contains $N_i$ samples. This metric is measured in HPatches~\cite{balntas2017hpatches} in three tasks: verification, retrieval and matching. In verification the descriptor is used to assign a score for a pair of patches indicating whether they correspond. In retrieval there is a query patch and the descriptor is used to retrieve the corresponding patch from a given set of patches which contains several positives as well as a big number of negatives. In matching, which is most relevant for the image matching pipeline, the task is to find the corresponding patch given a query patch and a set of patches which contains a single positive.

The second approach to evaluate a descriptor is more connected to its typical application. In IMW Phototourism (IMW PT)~\cite{Jin2020} the image matching pipeline, see Figure \ref{fig:cvprbench}, is used to estimate the relative pose between two cameras. Then the benchmark computes the angular error between the estimated and ground-truth rotation and translation vectors. A threshold is used to a binary result, indicating whether the estimate is good enough within the tolerance limit, and the average accuracy is calculated over all camera pairs. By integrating over a set of thresholds from 0\degree\ to X\degree, we obtain mean average accuracy mAA(5\degree) for X=5 and mAA(10\degree) for X=10.

In IMW PT~\cite{Jin2020} we use 8k Difference of Gaussians (DoG) keypoints, DEGENSAC~\cite{chum2005two} for model estimation and "both" matching strategy, i.e. with the requirement that matches are mutually nearest. FLANN library~\cite{muja2009fast} is used for approximate nearest neighbour search.

% is the most related to the usual real world use case. The evaluation pipeline was described in Section \ref{sec:introduction}. The benchmark has ground truth available between pairs of images from 3D reconstructions. It measures mean average accuracy (mAA) based on the error of the estimate of camera pose. The testing split contains 10 testing scenes that cover viewpoint changes and some illumination and season-related changes. We list the results in Table \ref{tab:IMW}. Our model achieves SOTA in ...

\chapter{Improving HardNet}
\section{Datasets}
The quality of the learned local descriptor depends on the data it is trained on. There are a variety of datasets~\cite{dturobot2012, RobotCarDatasetIJRR} of image sequences suitable for descriptor training, but only a few of them contain also keypoint correspondences, which are needed for the extraction of patches. In the current section we will describe existing datasets and the one we present.
%The datasets with such ground truth available differ mainly in the detector used for obtaining the feature points, noise applied before extraction of patches and the characteristics of the captured scenes. 
%There are only three relatively large-scale patch datasets for descriptor learning. Two of them: UBC Phototour and PS~\cite{cite-PS} cover mostly viewpoint changes, other datasets include also sequences with some amount of illumination change. 
%The most recent one is CVPR IMW dataset, which uses a large-scale 3D reconstruction of scenes from Phototourism data. It is intended as a part of an extensive benchmark for image matching methods but can be easily used for training as well. Our contribution is the AMOS Patches dataset, which is extracted from world-wide static webcams and is particularly useful for improving robustness to illumination and appearance changes.

\subsection{Existing Datasets}
\paragraph{UBC Phototour \cite{snavely2008modeling, goesele2007multi}}
UBC Phototour dataset\footnote{available at http://matthewalunbrown.com/patchdata/patchdata.html} (also called Brown) comprises 3 scenes: Liberty, Notredame and Yosemite. Point cloud 3D reconstructions and dense depth maps were constructed from original images with the help of Bundler~\cite{snavely2006photo} structure-from-motion library to obtain sets of corresponding patches. Each scene was reconstructed using Difference of Gaussian and Harris feature point detectors. Liberty sequence has been often been used as a single dataset for training\cite{mishchuk2017working}. It was used for testing as well, but with the rise of learned descriptors, the usually used FPR95 metric is already under 1\%, making this benchmark saturated.

\paragraph{HPatches \cite{brown2007automatic}}
This dataset\footnote{available at https://github.com/hpatches/hpatches-dataset} is a part of a benchmark for local feature descriptors. It contains 116 scenes, 6 images each, which are split to illumination (57 scenes) and viewpoint subset, based on the main nuisance factor. In the illumination split the scene is captured by a static webcam during a different time of the day and the viewpoint split contains significant camera pose changes. Images are related by a homography to the reference image. DoG, Hessian-Hessian and Harris-Laplace detectors were used to extract keypoints. The smallest patch size in the reference image is 16x16 pixels, while the extracted patch size is 65x65. Patches are further divided into difficulty levels (easy, hard and tough) according to the amount of synthetic noise added before projection. The applied transformation is composed of random rotation, translation, isotropic and anisotropic scaling.

\paragraph{The PhotoSynth Dataset \cite{mitra2018large}}
The PhotoSynth (PS) dataset\footnote{available at https://github.com/rmitra/PS-Dataset} contains patches from 30 scenes with illumination and viewpoint changes. COLMAP \cite{schonberger2016structure} SfM software was used to create 3D reconstructions with verified patch correspondences. Each patch is scale and rotation normalized. Minimum crop size is 20$\times$20 pixels, maximum is 128$\times$128 pixels. The cropped patches are resized to 48x48px. Around 400 000 patches were extracted from each scene. The trained descriptor achieved state-of-the-art performance at the time on the HPatches benchmark in the viewpoint split. It however sacrifices robustness to photometric changes.

\paragraph{IMW Phototourism 2020 \cite{Jin2020}}
This dataset\footnote{available at https://github.com/vcg-uvic/image-matching-benchmark} was recently created for the upcoming CVPR Image Matching Workshop and is a part of extensive evaluation software. Currently it contains 13 scenes for training, 3 for validation and 10 for testing. COLMAP \cite{schonberger2016structure} SfM software was used to create 3D reconstructions with verified patch correspondences. Each folder with a 3D reconstructed scene contains 2D to 3D point correspondences. It does not provide information about scale and rotation and the extracted patches cannot be easily normalized. However, we have observed that it is not a big issue in terms of the performance of the trained descriptor. Because the images are obtained from Phototourism data, it covers some changes in weather and lighting. The number of extracted patches is proportional to the total number of keypoints in all images and varies highly across the scenes. E.g. Brandenburg Gate contains over 2 million of patches, while Trevi Fountain contains over 15 million.

\subsection{AMOS Patches}
\label{sec:creating_patches}

\begin{figure}[htb]
\centering

\begin{subfigure}[c]{\linewidth}
  \frame{\includegraphics[width=\linewidth]{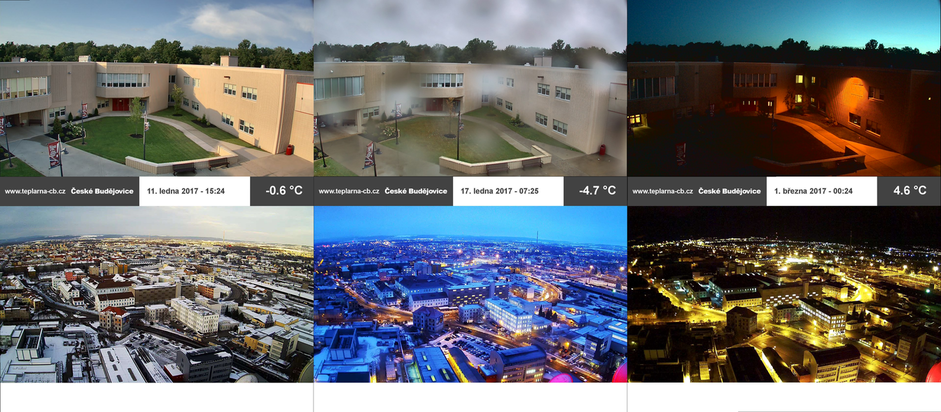}}
  \caption{}
\end{subfigure}\hfill\par
\begin{subfigure}[c]{\linewidth}
  \frame{\includegraphics[width=\linewidth]{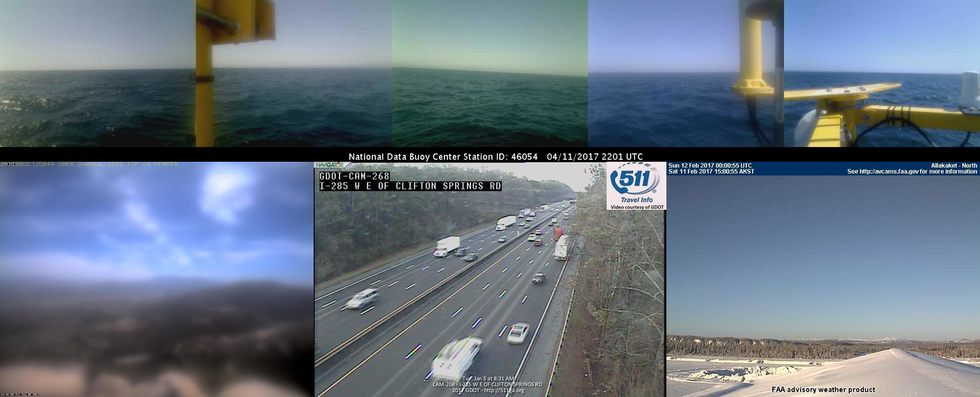}}
  \caption{}
\end{subfigure}\hfill\par

\caption[The AMOS dataset.]{The AMOS dataset \cite{jacobs2009global, jacobs2007consistent} - example images from (a) cameras contributing to the AMOS patches set and (b) cameras unsuitable for descriptor training because of blur, dynamic content or dominant sky. }
\label{fig:examples}
\end{figure}

\begin{figure}
\centering

\includegraphics[scale=0.16, width=\textwidth]{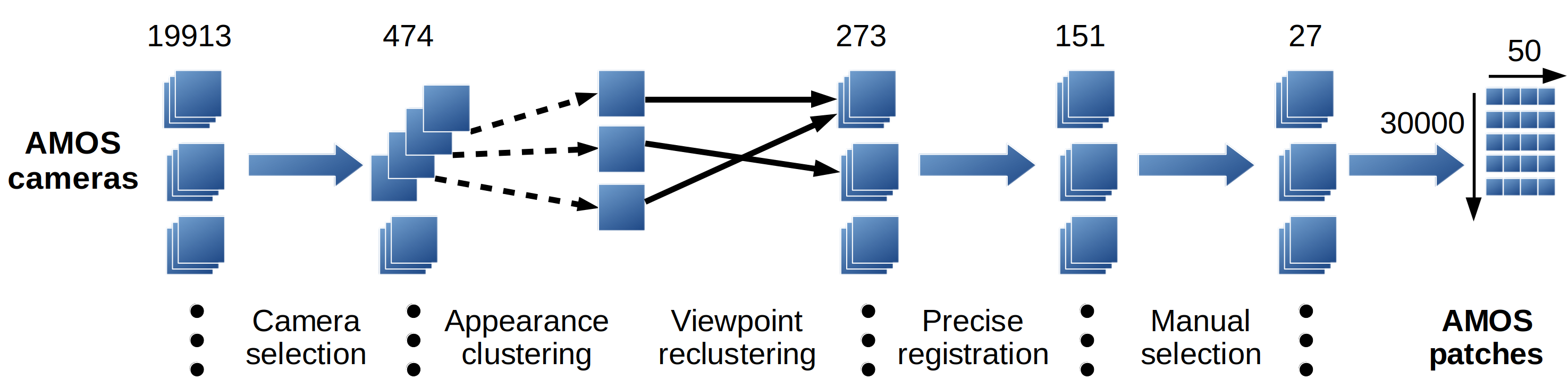}

\caption[The AMOS Patches extraction pipeline.]{The AMOS Patches extraction pipeline: camera selection to filter out dynamic or empty scenes, appearance clustering to remove redundant images, viewpoint reclustering to tackle switching cameras, precise registration for further filtering, manual pruning for the final selection of views and patch sampling. The numbers above denote the number of cameras in each stage.}
% \caption[The pipeline of AMOS Patches.]{The pipeline of AMOS patches consists of: camera selection to filter out dynamic or empty scenes, appearance clustering to remove redundant images, viewpoint reclustering to tackle switching cameras, precise registration for further filtering, manual pruning for the final selection of views and patch sampling. The numbers above denote the number of cameras in each stage.}
\label{fig:pipeline}
\end{figure}

There was little training data available for improving local descriptor robustness to photometric changes. We propose to fill this gap by introducing AMOS Patches dataset. It is created from the AMOS collection of world-wide static webcams and covers various illumination and season-related acquisition conditions.

AMOS~\cite{jacobs2009global, jacobs2007consistent} is a continuously growing publicly available dataset collected from outdoor webcams, currently containing over one billion (or 20 TB) images. It is organized into individual camera directories, which are split into folders according to the year and month of the acquisition. The size of the images varies, and so does their quality and the number of images in each camera directory. A typical AMOS camera is static and has approximately size of  300$\times$300 pixels. Many cameras store images in all seasons and during the whole day.

The advantage of static cameras lies in the fact that they show the same structure under different weather and lighting conditions. Therefore, if observing a static scene, they \emph{seem to be} highly suitable for training of local feature descriptor robust to illumination and appearance changes.

We learned the hard way that using this type of data is not trivial. Firstly, due to the dataset size, it is not feasible with moderate computing power to fit such data into memory. Moreover, preprocessing would take a prohibitive amount of time. Secondly, the training procedure is sensitive to the misregistration of the images and the presence and size of moving objects. Many cameras experience technical issues such as: being out of focus, rotating over time, displaying highly dynamic scene (e.g. sky, sea waves), which all significantly hurt the performance of the trained descriptor, as discussed later.
 
Therefore, we developed a pipeline for the creation of AMOS Patches, shown in Figure~\ref{fig:pipeline}, which entails several steps to create a clean dataset with veridical patch correspondences. These methods focus on the selection of cameras and images, detection of view switching in a camera and the registration of images. Because of several not easily detectable problems, it was still necessary to perform a final manual check of the selected image sets.

\paragraph{Camera selection}
The first step --- camera selection --- aims at choosing a subset of cameras that are suitable for training, i.e.\ do not produce very dark images, are sharp and do not display moving objects like cars or boats.

The procedure uses two neural networks, a sky detector~\cite{mihail2016sky} and an object detector \cite{torchcv}, and computes simple statistics for each of 20 randomly chosen images in each camera directory.

The camera selection took approximately one week on a single PC (Intel Xeon CPU E5-2620) with one GPU GTX Titan X. Processing more images by the neural network detectors increases both the precision of the method and the running time. Our choice is therefore based on the (limited) available computation power.

Each image is then checked whether it satisfies the following conditions:

\begin{itemize}
\item $f_{1}~:$ sky area $<50\%$ \hfill \textit{not empty}
\item $f_{2}~:$ no detected cars or boats \hfill \textit{not dynamic}
\item $f_{3}~:$ $\mathrm{Var}(\nabla^{2}$ \text{image}) $\geq 180$ \hfill \textit{sharp}
\item $f_{4}:$ mean pixel intensity $>30$  \hfill \textit{not black}
\item $f_{5}:$ image size $ >(700,700)$  \hfill \textit{large}
\end{itemize}

A camera is kept if at least 14 out of the 20 images pass the check.

The filter $f_5$ is the most restrictive, it removes $91\%$ of the cameras -- AMOS contains mostly low-resolution images. The reasoning behind using $f_5$ is that images of smaller size often observe a motorway or are blurred. Also, such cameras would not generate many patches. We want to select only a relatively small subset of the cameras with the predefined characteristics and therefore an incorrect removal of a camera is not a problem.

Several cameras were removed because of corrupted image files. The resulting set contains 474 camera folders which were subject to subsequent preprocessing.

\paragraph{Appearance clustering by K-means}
The resulting data is of sufficient quality, but it is highly redundant: images shot in 10 minute intervals are often indistinguishable and very common. To select sufficiently diverse image sets, we run the K-means clustering algorithm with $K$=120 to keep the most representative images. We use the fc6 layer of the ImageNet-pretrained AlexNet~\cite{krizhevsky2012imagenet} network as the global image descriptor. While not being the state-of-the-art, AlexNet is still the most effective architecture in terms of speed~\cite{Canziani2016}, with acceptable quality.

At this stage of the pipeline, there are $K$=120 images for each of the $C$=474 cameras selected, a feasible number for training with the computational resources available.

Feature descriptor training with patches selected from this image set was not successful.
We were unable to achieve accuracy higher than 49.1 mean average precision (mAP) in the HPatches matching task; 
the state-of-the-art is 59.1 mAP -- GeoDesc~\cite{geodesc2018}.

\paragraph{Viewpoint clustering with MODS}
After examining the data closely, we found that many of the cameras switch between a few views, which breaks our assumption for the generation of ground truth correspondences via identity transformation. In order to filter out the non-matching views, we run MODS~\cite{mishkin2015mods}, a fast method for two-view matching, and split each camera folder into clusters, called views, by applying a threshold on the number of inliers and the difference between the homography matrix and the identity transform.

Let $(x_1, x_2, ..., x_K)$ be a set of images in a camera folder in arbitrary order. MODS matching is first run on pairs $(x_1,x_2),(x_1,x_3),...(x_1,x_K)$. Image $x_1$ becomes the reference image in a newly created view, which contains $x_i$ for which the registration yields more than 50 inliers and SAD($H(x_1, x_i), I_3) < 50$. SAD denotes the sum of absolute differences, $H$ denotes a homography matrix normalized by the element in position $(3,3)$, $I_3$ is 3x3 identity matrix. All images in the created view are then removed from the processed image set. The step is repeated until no images remain.

We observed that the number of the resulting views in one camera folder depends on phenomena other than camera movement. For example, in cases where there is a fog or very rainy weather, MODS fails to match most of the image pairs and many of them form a single element cluster, which is excluded from further processing. For each camera, we keep only the view with the largest number of images, if it has more than 50. Each remaining view is reduced to 50 images by random selection.

\begin{figure}[htb]
\centering

\includegraphics[max width=1\textwidth]{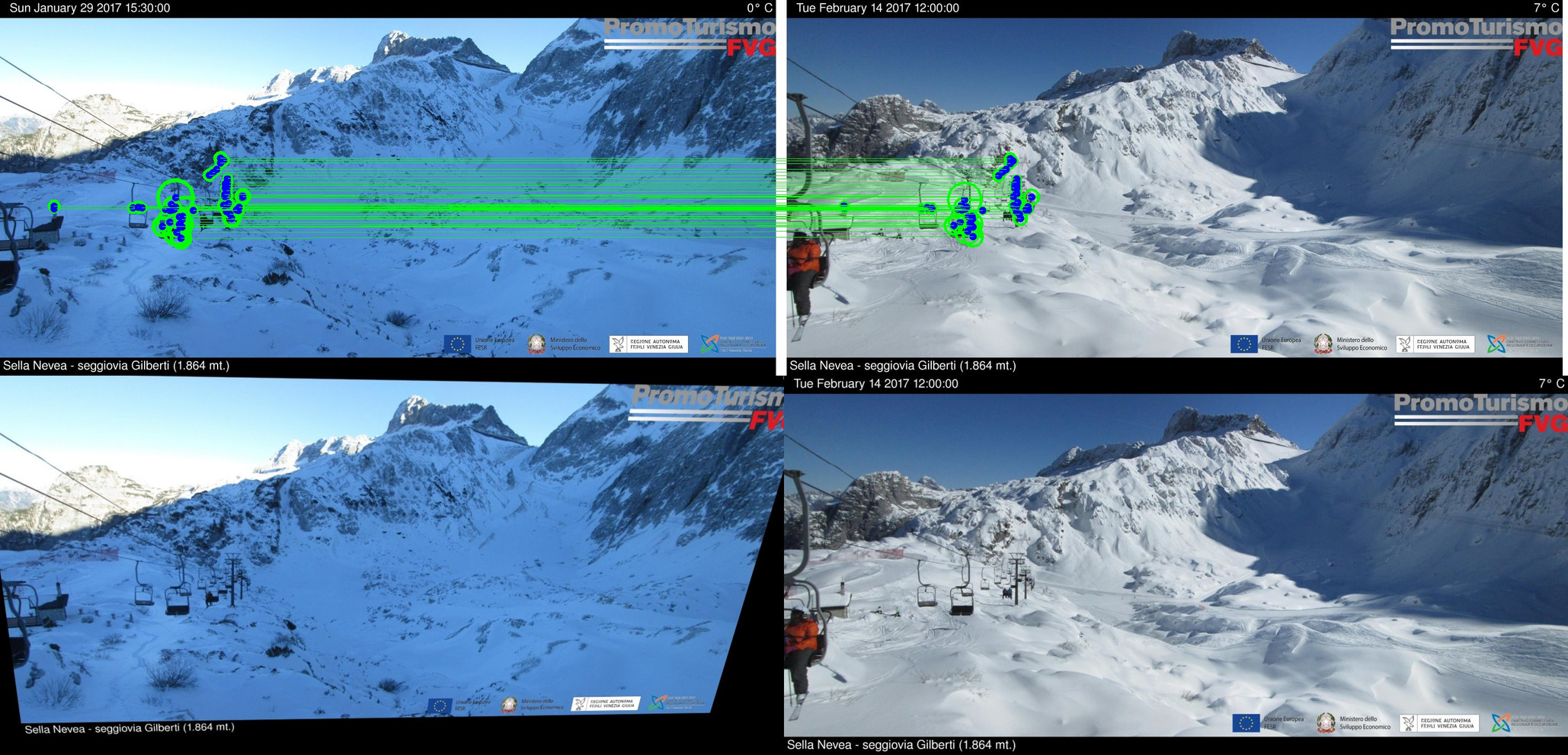}

\caption[MODS registration failure example.]{MODS registration failure, most of the correspondences are on moving structures.
 Top: an image pair with marked inliers. Bottom: misregistered image (left) and the reference.}
\label{fig:mods_wrong}
\end{figure}

\paragraph{Registration with GDB-ICP}
While the MODS method is effective in matching and subsequent reclustering of camera sequences, in most cases the estimate of the global homography is not sufficiently precise. MODS often outputs a homography valid for only a small area in the image, see the example is shown in Figure~\ref{fig:mods_wrong}. Therefore, the views contain also images that are not correctly aligned. To alleviate the problem, we run Generalized Dual Bootstrap-ICP \cite{yang2007registration} to prune the set of views, keeping those where this second registration is successful.

The registration proceeds as follows. Each view folder contains images $(x_1, x_2, ..., x_{50})$, where image $x_1$ is the MODS reference. The GDB-ICP registration is run on pairs $(x_1, x_2), (x_1, x_3),... (x_1, x_{50})$ and warped images $x'_2, x'_3, ..., x'_{50}$ are obtained. If registration fails on any pair, the whole view is removed.

After the precise registration with GDB-ICP, 151 views remained. 
It is feasible to manually inspect such a set.

\begin{figure}[htb]
\centering

\includegraphics[max width=1\textwidth]{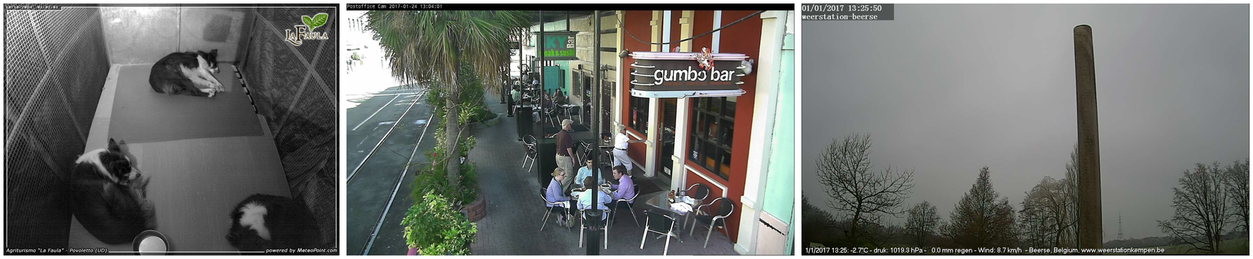}

\caption[AMOS dataset manually pruned views.]{AMOS dataset manually pruned views. Examples of dynamic scenes (left, center) and a cloud-dominated scene not removed during camera selection (right). }
\label{fig:manual_images}
\end{figure}

\paragraph{Manual Pruning}
A few problems remain, see Figure \ref{fig:manual_images}, such as dynamic scenes,
undetected sky (the sky detector fires mostly on the clear blue sky). As a precaution, we also removed views with very similar content and views from different cameras observing the same place from a different viewpoint. 
We tried to use the scene segmentation network \cite{zhou2018semantic} to detect moving objects, but the result was not satisfactory. The final selection is therefore done by hand, resulting in a set of 27 folders with 50 images each, which we call AMOS Views. In this step we also select another 7 views which form a test set in our AMOS Patches matching benchmark.

\subsubsection{Extracting AMOS Patches}
\label{sec:extracting}
The last phase of the AMOS Patches pipeline consists of sampling images to obtain patch centers, scales and angles, and subsequent cropping from source images. Here we explore two approaches. In the first one we select patch centers and scales randomly based on a probability map. The probability map construction is discussed below. In the second approach we deterministically pick local maxima in the Hessian detector response map. In both cases we use each selected point to extract a set of corresponding patches of size 96$\times$96 pixels. Each patch set contains 50 corresponding patches, i.e. one per each image in a view. Here we assume the images in each view are well registered, meaning that a mapping between image coordinates is (almost) identity. The assumption holds due to previous parts of the pipeline. The number of patch sets is manually set in the first approach, and semi-automatically determined in the second approach.

In the following experiments we train for 10 "epochs", 5 million pairs of samples each, with batch size = 1024, learning rate = 20, SGD optimizer with momentum $= 0.9$ and a linear learning rate decay schedule unless stated otherwise. We use "epoch" not in a conventional sense as "going through all examples in the dataset," but in the sense of "going through all generated pairs."

% During training, we apply random affine transformation and cropping to get patches of smaller size. First, random rotation from range $(\ang{-25},\ang{25})$, scaling from range $(0.8, 1.4)$ and shear are applied. Second, we crop a 64 times 64 center of a patch. Then we crop a random area with scale in $(0.7,1.0)$ and aspect ratio in $(0.9,1.1)$ w.r.t. to the input patch and the patch is resized to 32 times 32 pixels. These transformed patches are the input for training. We train for 20 epochs with batch size of 1024, learning rate = 20, SGD optimizer with momentum $= 0.9$ and use a linear learning rate decay schedule.

% older:
% During training, we apply random affine transformation and cropping to get patches of smaller size. First, random rotation from range $(\ang{-25},\ang{25})$, scaling from range $(0.8, 1.4)$ and shear are applied. Second, from a 64 times 64 center of a patch we crop a 32 times 32 region with a random scale. These transformed patches are the input for training.

\subsubsection{Sampling of Feature Points}
\label{sec:points_random}
Here we explore a sampling of feature points based on a probability map. There are many ways how to construct the probability distribution. First, it relies on the choice of a response function which is evaluated for each position in an input image. Secondly, it depends on the choice of input images for the response function and how we average the obtained results. In the following text we explain these choices in more detail and find the best setting. Once the random distribution is computed, scales and angles are chosen randomly and the patches are extracted.

\paragraph{Response Function}
First we decide on the response function. In Table \ref{tab:response} we compare several possibilities. The differences in mAP scores are not large. The best mAP is achieved by the square root of the determinant of the Hessian matrix. If we do not extract the root, the result is slightly worse. The worst results are achieved by the uniform distribution together with the square root of the determinant of the Hessian matrix and nonmax suppression (NMS). We choose the $\textstyle\sqrt{\text{Hessian}}$ as the optimal response function. In the next step we explore two ways how we can perform the sampling procedure: by averaging before or after computing the response.

\begin{table}[htb]
\centering
\caption[Patch sampling: Influence of the response function on HPatches matching score (mAP).]{Patch sampling: Influence of the response function on HPatches matching score (mAP). By "Hessian" we denote shortly the determinant of the Hessian matrix.}

\begin{tabular}{lc}
\toprule
Weighting  & mAP\\
\midrule

Uniform  & 56.20 \\
Hessian   & 56.39 \\
$\textstyle\sqrt{\text{Hessian}}$& \textbf{56.49} \\
NMS($ \textstyle\sqrt{\text{Hessian}}$)  & 56.18 \\
\bottomrule
\end{tabular}
\label{tab:response}
   
\end{table}

\paragraph{Representative Image}
The first way of sampling of the patches comprises acquiring a representative image, on which we subsequently evaluate the chosen response function - the output is then normalized and is used as the random distribution. We compare three possibilities: random, mean and median, see Figure \ref{tab:source_img}. "Random" refers to choosing a single image from the dataset. In "mean" and "median" we apply the corresponding averaging over all images in a view. One can see that all three choices are comparable. "mean" averaging gives the best result by a small margin.

\begin{table}[htb]
\centering
\caption[Patch sampling: Influence of the image averaging on HPatches matching score (mAP).]{Patch sampling: Influence of the image averaging on HPatches matching score (mAP). Weighting function is $ \textstyle\sqrt{\text{Hessian}}$.}

\begin{tabular}{lc}
\toprule
Image & mAP\\
\midrule
random & 56.49 \\
median  & 56.44 \\
mean & \textbf{56.58} \\
\bottomrule

\end{tabular}
\label{tab:source_img}
\end{table}

\paragraph{Averaging the Responses}
In this second way of sampling, we first evaluate the response function for all images in a view. The random distribution is then acquired by averaging all the responses or by choosing a single response (denoted as none), see Figure \ref{tab:averaging}. The latter variant achieves the highest mAP score. It does not beat the best choice in the previous paragraph.

\begin{table}[htb]
\centering
\caption[Patch sampling: Influence of the response averaging on HPatches matching score (mAP).]{Patch sampling: Influence of the response averaging on HPatches matching score (mAP). Weighting function is $ \textstyle\sqrt{\text{Hessian}}$.}
   
\begin{tabular}{lc}
\toprule
Averaging & mAP\\
\midrule

none & \textbf{56.49} \\
mean  & 56.10 \\
median & 56.45 \\
\bottomrule

\end{tabular}
\label{tab:averaging}
\end{table}

Overall, we conclude that the process of construction of a probability map is not as important as one may think. Although the images contain a lot of noise, choosing a random representative image is just as good as some of the more robust approaches.

% Second, one may evaluate the response function over all images in a view and average the outputs. The resulting 2D map is then used as a probability mask for the selection of patch centers. Scales and angles are sampled independently at random from a predefined range. 

\begin{figure}[htb]
\centering

\includegraphics[max width=0.5\textwidth]{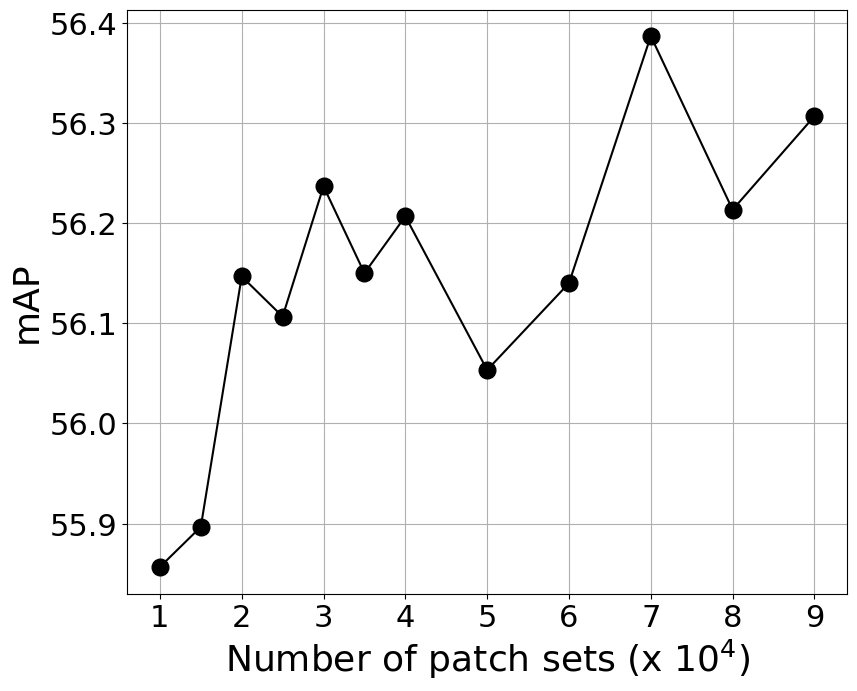}

\caption[HardNet mAP score in HPatches matching task evaluated for different sizes of AMOS patches training dataset]{HardNet mAP score in HPatches matching task evaluated for different sizes of AMOS patches training dataset. Each value is an average over 3 different randomly generated datasets of the same size.}
\label{fig:n_patches}
\end{figure}

\paragraph{Number of Patch Sets}
We have determined the optimal way of sampling patches using a probability distribution, but we do not know what should be the size of the dataset. For this reason we created several datasets containing a different number of patch sets, from 10 000 to 90 000. Figure~\ref{fig:n_patches} shows that the performance of the trained descriptor roughly increases with the dataset size. We select extracting around 30 000 patch sets as a reasonable trade-off between performance and dataset size.

% Two evaluation tasks are considered. In the matching task, there are two equally sized sets of patches from two different images. The descriptor is used to find a bijection between them. The `average precision (AP) over discrete recall levels is evaluated for each such pair of images. Averaging the results over a number of image pairs gives mAP (mean AP). In the verification task there is a set of pairs of patches. The descriptor assigns a score that the two patches in a pair correspond. The precision-recall curve is then plotted based on the sorted (according to the score) list of patch pairs distances.

\subsubsection{Deterministic Selection of Feature Points}
\label{sec:points_deter}

\begin{figure}[htb]
\centering

\includegraphics[width=0.99\linewidth]{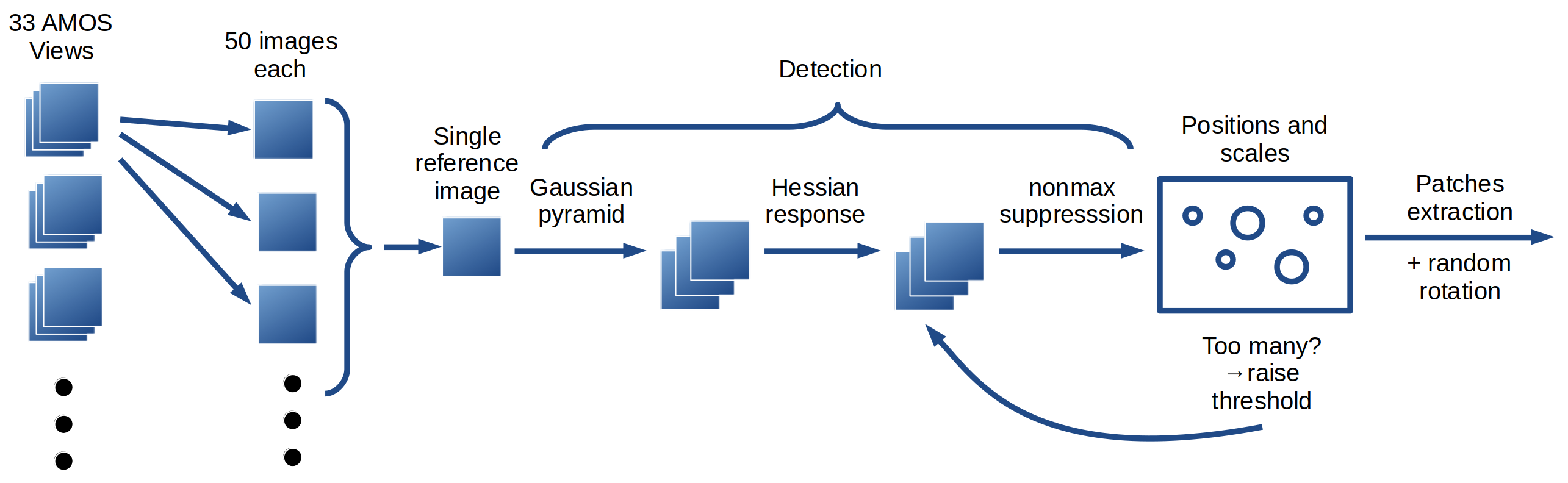}

\caption[Deterministic selection of feature points for AMOS Patches.]{Part of the AMOS Patches extraction pipeline \cite{HardNetAMOS2019}, which selects feature points in a deterministic way. Here we consider we already have filtered and registered set of camera views as input. The reference image is per-pixel median over the images. Detection part is fully described in section \ref{sec:points_deter}. In the end of the pipeline, patches are extracted from all 50 images using the same obtained set of positions and scales.}
% \caption[Deterministic selection of feature points.]{Depicted is part of the adjusted AMOS Patches extraction pipeline \cite{HardNetAMOS2019}. Here we consider we already have filtered and registered set of camera views as input. The reference image is per-pixel median over the images. Detection part is fully described in section \ref{sec:points_deter}. In the end of the pipeline, patches are extracted from all 50 images using the same obtained set of positions and scales.}
\label{fig:flow}
\end{figure}

Here we describe the second approach to extracting AMOS Patches. We noticed the previous method is somewhat distant from the common implementation in image matching, where feature points typically represent local maxima in the output of a response function. This approach is therefore very similar to the Hessian detector. One difference is that we do not use nonmax suppression in a 3D scalespace, but instead apply it for each scale in the image pyramid separately.

The input is a representative (e.g. median) image of a single AMOS View. From this input, a sequence of blurred images is created using incrementally larger Gaussian filter. Let $\sigma^{k_1},\sigma^{k_1 +1},...,\sigma^{k_2}, k_i \in \mathbb{N}_0, \sigma=1.8$ empirically set, be chosen sequence of standard deviations of the normal distribution and the levels be indexed $k_1,k_1 +1, ...,k_2$ according to the assigned sigma. The Gaussian kernel has width 6$\sigma^{k_i}$ in each level $k_i$ and is convolved over the image. Output of this step is a sequence of images.

In the next step, Hessian kernel is convolved over all images, the response output is multiplied by $1 / \sigma^{4k_i}$ in each level and then maxpooling is applied with kernel size equal to $2\sigma^{k_i} \cdot bs$, where $bs$ denotes base scale and is usually around 16 (half of the patch size). Every local maxima under empirically chosen threshold = $0.00016$ is discarded so that abundance of feature points is still generated per image. If too many feature points are generated, the threshold is multiplied by $2$ and the algorithm starts over. We decided to keep at most 2000 feature points per input image (i.e. per single view).

The output of the described procedure is a set of locations in the representative image together with scales. Angles are chosen randomly for each patch set. As in the previous approach, the last step is the extraction of patches.

% In the adjusted implementation of the detector in AMOS Patches pipeline, $k_1, k_2, bs$, are the parameters which we search for in the ablation study, tab \ref{tab:compare}. We pick $k_1=0, k_2=5, bs=30$, corresponding to min and max patch scales $s_{min}=30, s_{max}=315$, as the optimal setting for which the mean of HPatches full mAP and AMOS test mAP scores are the highest. Notice that the minimal scale $s_{min}$ was equal to 16 in the original probability map based variant of the detector.

\paragraph{Hyperparameters}

\begin{table*}[htb]
\centering
\caption[Hyperparameter search for a deterministic selection of patches.]{HPatches and AMOS Patches matching scores (mAP). $s_{min}, s_{max}$ denote the smallest and largest patch scale, which is computed as $s_{min}=bs\cdot \sigma^{k_1}, s_{max}=bs\cdot \sigma^{k_2}, \sigma=1.8$. $bs$ denotes the base scale, which is also proportional to the kernel size of maxpool operator in detection of points of each scale.}

\begin{tabular}{lccccc}
\toprule
$s_{min}$ & $s_{max}$ &\multicolumn{3}{c}{HPatches subset} & AMOS Patches\\
& &illum & view & full  \\
\midrule

\multicolumn{4}{l}{HardNet} && \\
24 & 140 & 54.58 & 59.10 & 56.88 & 45.94 \\
24 & 252 & 54.28 & 58.68 & 56.52 & 46.20 \\
43 & 140 & 55.00 & 60.04 & 57.56 & 45.38 \\
43 & 252 & 54.97 & 60.09 & 57.57 & 45.32 \\
30 & 175 & \textbf{55.32} & \textbf{60.16} & \textbf{57.78} & \textbf{46.43} \\
30 & 315 & 55.01 & 60.02 & 57.56 & \textbf{46.43} \\
54 & 175 & 53.98 & 58.81 & 56.43 & 44.24 \\
54 & 315 & 53.51 & 58.60 & 56.10 & 44.93 \\
\midrule

\multicolumn{4}{l}{HardNet8} && \\
24 & 140 & 57.69 & 62.61 & 60.19 & 47.37 \\
24 & 252 & 57.73 & 62.62 & 60.22 & 47.47 \\
43 & 140 & 58.02 & 63.76 & 60.94 & 46.33 \\
43 & 252 & 57.99 & 63.48 & 60.78 & 46.26 \\
30 & 175 & \textbf{58.42} & \textbf{63.93} & \textbf{61.22} & 47.60 \\
30 & 315 & 58.27 & 63.69 & 61.02 & \textbf{47.75} \\
54 & 175 & 55.79 & 61.06 & 58.47 & 44.87 \\
54 & 315 & 55.84 & 61.06 & 58.49 & 45.11 \\
\bottomrule
\end{tabular}

\label{tab:compare}
\end{table*}

% \begin{table*}[htb]
% \centering
% \caption{HPatches and AMOS test matching scores (mAP). $s_{min}, s_{max}$ denote the smallest and largest patch scale, which is computed as $s_{min}=bs\cdot \sigma^{k_1}, s_{max}=bs\cdot \sigma^{k_2}, \sigma=1.8$. $bs$ denotes the base scale, which is also proportional to the kernel size of maxpool operator in detection of points of each scale. }

% \begin{tabular}{lccccc}
% \toprule
% $s_{min}$ & $s_{max}$ &\multicolumn{3}{c}{HPatches subset} & AMOS test\\
% & &illum & view & full  \\
% \midrule

% \multicolumn{4}{l}{HardNet} && \\
% 24 & 140 & 		55.82&59.44&	57.66&	61.07 \\
%  24 & 252 & 	55.71&59.36	&57.57&		60.96 \\
%  43 & 140 & \textbf{56.23}&\textbf{61.34}&\textbf{58.83}&59.82 \\
%  43 & 252 & 		55.79&60.32	&58.1&	59.76 \\
%  30 & 175 & 	56.09&60.27	&58.21&		60.96 \\
%  30 & 315 & \textbf{56.23}&60.07&58.19&\textbf{61.09} \\
%  54 & 175 & 		54.94&59.1	&57.05&	59.6 \\
%  54 & 315 & 	54.8&59.14	&57.01&		59.44 \\
% \midrule

% \multicolumn{4}{l}{HardNet-bigger} && \\
%  24 & 140 & 		59.29&63.78	&61.57&	62.71 \\
%  24 & 252 & 		59.24&63.51	&61.41&	62.69 \\
%  43 & 140 & 		59.25&64.01	&61.67&	61.34 \\
%  43 & 252 & 		58.96&63.7	&61.37&	61.41 \\
%  30 & 175 & 	59.68&64.09	&61.92	&	61.76 \\
%  30 & 315 &	\textbf{59.97}&\textbf{64.57}& \textbf{62.31}&\textbf{62.73} \\
%  54 & 175 & 		57.78&62.02	&59.94&	60.82 \\
%  54 & 315 & 	57.45&61.94	&59.73&		60.51 \\
% \bottomrule
% \end{tabular}

% \label{tab:compare}
% \end{table*}

In our implementation of the Hessian detector in AMOS Patches pipeline, $k_1, k_2, bs$, are the parameters which we search for in the ablation study, see Table \ref{tab:compare}. We pick $k_1=0, k_2=4, bs=30$, corresponding to min and max patch scales $s_{min}=30, s_{max}=175$, as the optimal setting for which the mean of HPatches full mAP and AMOS test mAP scores are the highest. With this setting we outperform the previous, probability-based, approach to sampling.

\subsubsection{AMOS Patches Test Set} In addition to the training data, we also create AMOS Patches test set, which is extracted from 7 pruned views, 10 images each. Keypoints are obtained using the SIFT detector. We consider the matching task and follow the HPatches protocol. 

% \subsubsection{AMOS Patches Test Set} In addition to the training data, we also create AMOS Patches test set, which is extracted from 7 pruned views. Keypoints are obtained using the SIFT detector. We consider the matching task, which is implemented similarly as in HPatches. During testing we take patches from pairs of images and use the inverse of the Euclidean distance between descriptors of corresponding patches as similarity score. Average precision is computed for all pairs over 10 selected images per each view. Mean average precision is obtained by taking mean over all the image pairs.

% Matching task is applied in both benchmarks. First, patch pairs are sorted with respect to the assigned similarity score. Then average precision is computed and by taking mean over all queries we obtain mean average precision (mAP). In the case of AMOS Patches the queries are constructed by considering all pairs over 10 images in each view.
% , which cover significant photometric and season related changes

\subsubsection{Sampling of Patches for a Batch}
In this section we test different approaches to sampling of patches for a batch. The composition of a batch usually has an impact on the efficiency of the training procedure, because of the averaging of the gradients over the samples. We use the hard-in-batch structured loss function, where the composition furthermore has an influence on the distribution of the negative patches. Therefore, one may expect that the way of sampling to be important in our training. In this section we use probability map-based extraction of patches to create a dataset with 30 000 patch sets.

\begin{figure}[htb]
\centering

\includegraphics[max width=0.5\textwidth]{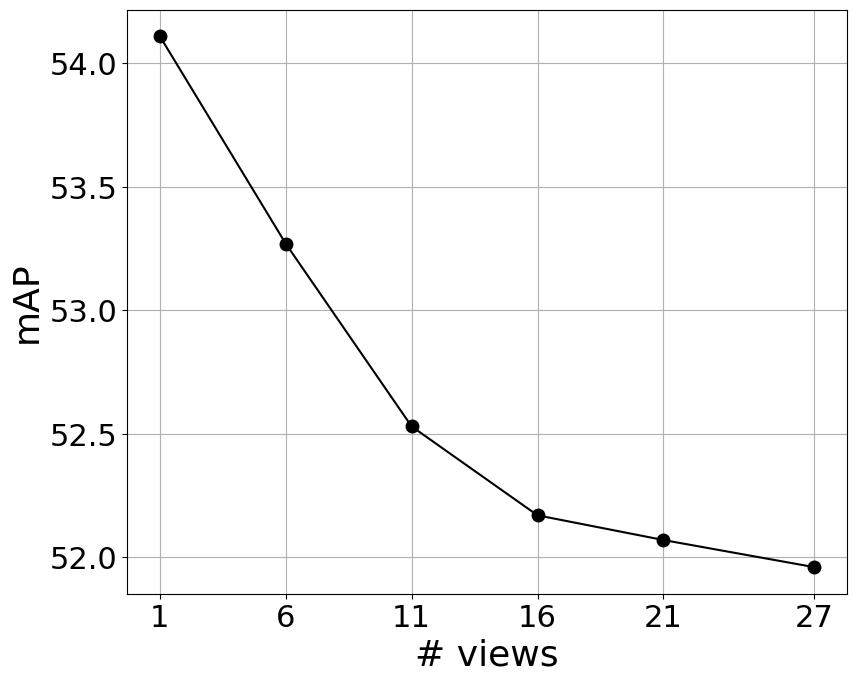}

\caption[HardNet mAP score in HPatches matching task as a function of the number of source views for a batch.]{HardNet mAP score in HPatches matching task as a function of the number of source views for a batch. Views are selected randomly in each iteration. Dataset consists of 27 views in total.}
\label{fig:n_cams}
\end{figure}

\paragraph{Uniform Sampling}
In this way of sampling we select $B$ patch sets for each batch. From each set we pick randomly two positive patches. Both selections are done using the uniform distribution. We train for 20 epochs with batch size = 256. The training dataset consists of AMOS Patches from 27 views in total. In Figure \ref{fig:n_cams} we show the performance (mAP) as a function of the number of views we select from for each batch. Notice the monotone trend, which we interpret as follows. Reducing the number of views increases the number of negative patches from the same scene, which are often the most difficult to distinguish.

If we increase the batch size to 1024 and choose the number of source views = 6, then the model achieves 56.57 mAP on HPatches full split. Note that we did not use this batch size in Figure \ref{fig:n_cams}, because it would be higher than the minimal number of patch sets there is in a view, i.e. the trend would be obscured for \# views = 1.

A related experiment was performed on the Liberty dataset, see Table \ref{tab:fewimgs}. An improvement is observed on AMOS Patches and IMW PT if patches are sampled from as few source images as possible. For this method we used publicly available file "interest.txt" containing IDs of reference images, which have been usually omitted during training.

\begin{table}[htb]
\centering
\caption[Evaluation of a method of sampling patches for a batch.]{Evaluation of a method of sampling patches for a batch. Either any or as few as possible source images are selected. Mean average precision (mAP) is measured on HPatches and AMOS Patches and mean average accuracy mAA(10$\degree$) on the CVPR IMW benchmark.}

\begin{tabular}{lccc}
\toprule
Source images & HPatches & AMOS Patches & IMW PT \\
\midrule
Any & \textbf{52.96} & 44.21 & 68.17 \\
As few as possible & 51.53 & \textbf{45.96} & \textbf{68.63} \\
\bottomrule

\end{tabular}
\label{tab:fewimgs}
\end{table}

\paragraph{Sampling from Pairs of Images}

\begin{figure*}[htb]
\centering

\includegraphics[width=0.45\linewidth]{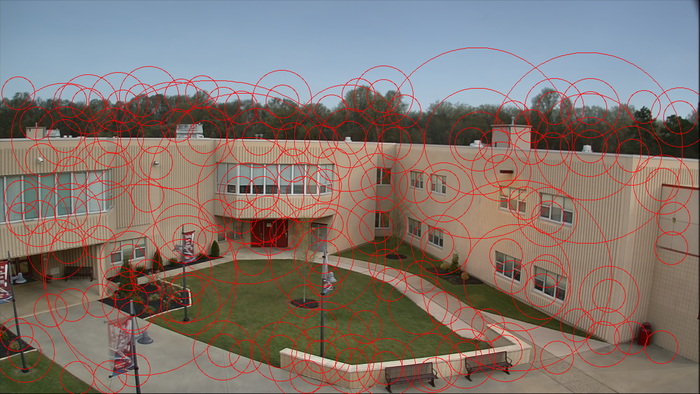}
\includegraphics[width=0.45\linewidth]{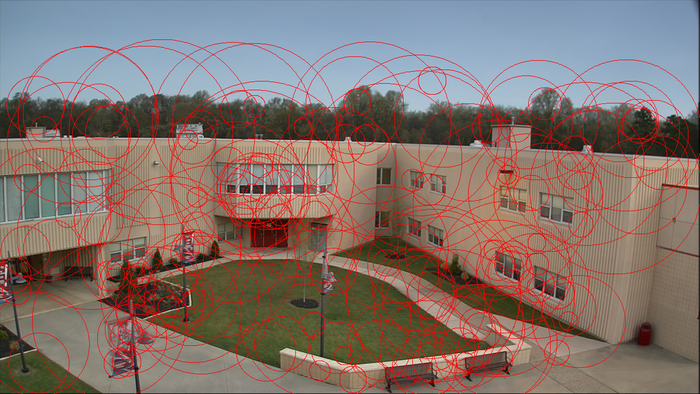}

\caption[Comparison of two approaches to sampling of patches.]{Two approaches to sampling of patches. Left: using a probability map, right: using multiscale Hessian detector with scale-wise nonmaxima suppresion. Both variants contain 186 patches. Patch centers and scales marked on single AMOS Views image.}
% \caption[Detected patch centers and scales using Hessian detector.]{Detected patch centers and scales using Hessian detector marked on single AMOS Views image. Left: probability map-based sampling, right: multiscale Hessian detector-based sampling with scale-wise nonmaxima suppresion. Both variants contain 186 patches.}

%original implementation \cite{HardNetAMOS2019} using probability map, right: changed implementation using multiscale detector with scale-wise nonmaxima suppresion. Both variants contain 186 patches.}
\label{fig:samples}
\end{figure*}

The previous approach is somewhat counter-intuitive since patches from many images from a single view may co-appear in a batch, creating a situation that does not occur in stereo matching. In the new strategy of sampling we run sequentially over all camera views until the batch is filled, selecting patches always only from a pair of images - and the selection of the pair is random so that all patches have equal chance to contribute. HardNet trained with such sampling approach achieves 56.42 mAP on HPatches. Therefore, contrary to our intuition, this approach is comparable to uniform sampling.

\begin{figure}[htb]
\centering

\includegraphics[width=\linewidth]{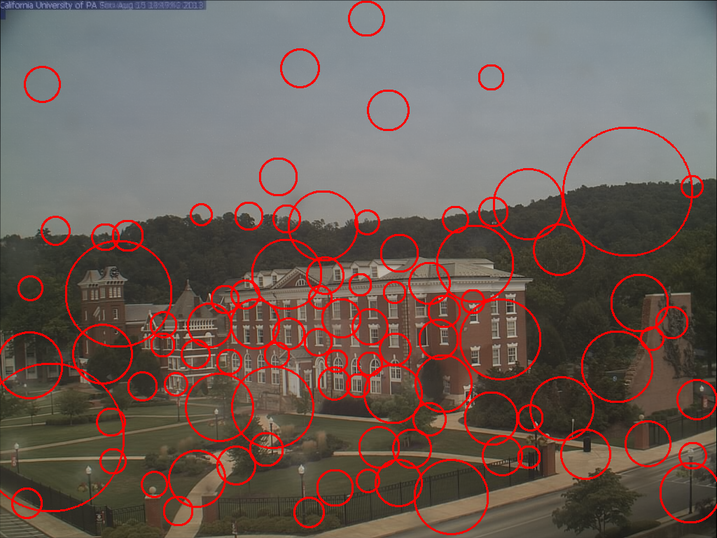}

\caption[Visualization of a strategy of sampling of pairs of patches enforcing patch overlap to be small.]{A strategy of sampling of pairs for a batch, in which we enforce patch overlap across the same scale to be small.}
% \caption[Nonoverlapping way of sampling]{A strategy of sampling of pairs for a batch, in which we enforce patch overlap across the same scale to be small.}
\label{fig:colls}
\end{figure}

\paragraph{Avoiding Collisions}
Another, in a sense orthogonal, approach to sample patches for a batch uses information about overlaps between patches generated from a camera view. During the sampling step, after selecting a sample from the pool of available patches, the pool is iterated and overlapping patches are discarded, see Figure \ref{fig:colls}. The idea of this method comes from the "first geometrically inconsistent (GI)"~\cite{mishkin2015mods} criterion, which is used in image matching, where it applies a threshold on the ratio between the distance to the first and second keypoints, which considers a ratio between descriptors of patches, which are geometrically inconsistent, i.e. not overlapping. Our idea was to provide only such patches as negatives in the triplet loss.

The trained model achieves 55.70 mAP on HPatches and so this approach is inferior to the previous ones. Our hypothesis is that it may be due to preference of smaller -- therefore containing less information -- patches during sampling. Also, in real stereo matching patches do overlap (see Figure \ref{fig:samples}, right).

\subsection{Experiments with AMOS Patches}
\label{sec:expsamos}
\begin{figure}[htb]
\centering

\includegraphics[max width=0.5\textwidth]{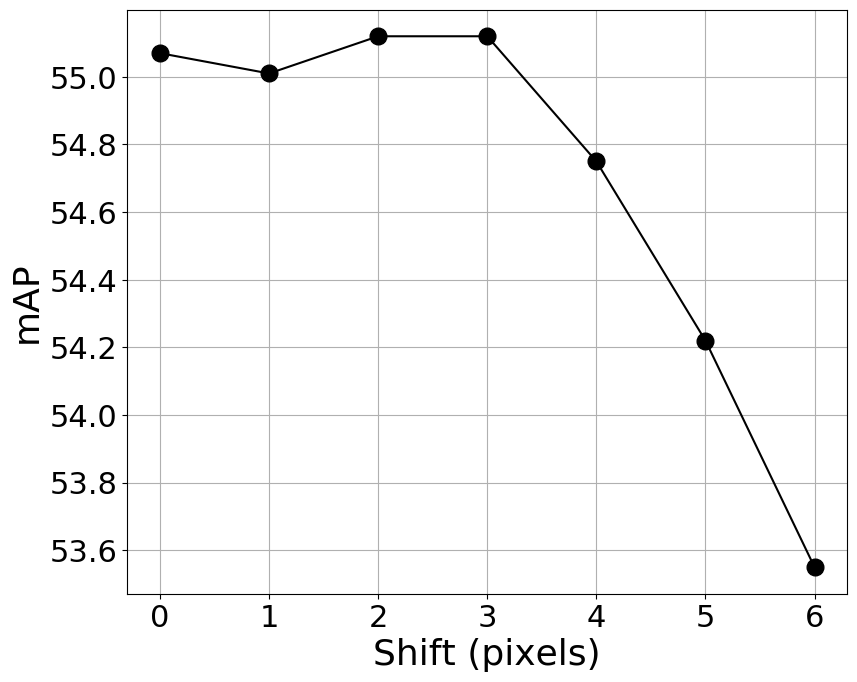}

\caption[The mAP score of HardNet trained on AMOS patches displaced by different shifts.]{HPatches matching. The mAP score of HardNet trained on AMOS patches displaced by different shifts.}
\label{fig:deregistration}
\end{figure}

\paragraph{The Importance of View Registration}
In this experiment we show the importance of the precise alignment of images. We displace each patch by different shifts and observe the influence on the HPatches matching score, see Figure \ref{fig:deregistration}. Note how the performance of the descriptor improves with a small shift, but then quickly deteriorates. We use the patch selection approach using probability map, $\#$source views $=27$ (all), 30000 patch sets and Hessian weighting without averaging.

\paragraph{Data Modality}
To the best of our knowledge, learned descriptors have usually been trained on grayscale data -- Brown, PS and HPatches dataset contain only such. Here we experiment with other modalities. We extract RGB patches from AMOS Views and compare two architectures: HardNet8 and HardNet8 with learned color conversion. The latter model has prepended two more convolutional layers with kernel size = 1 which have 10 filters (first layer) and 1 filter (second layer) similar to~\cite{Systematic2017}. We can see in Table \ref{tab:modality} that this model improves the mAP score on the AMOS Patches test set. The difference is however not large and using RGB data has its drawbacks, for example bigger size of the dataset or incompatibility with other benchmarks (HPatches, UBC Phototour). We believe the improvement is not significant enough to compensate the demerits. 

We also tried to make use of MegaDepth~\cite{MDLi18}, a tool for single-view depth prediction, to enrich the modality our dataset. MegaDepth pretrained model was applied to each image in our AMOS Views dataset. The estimated depth maps were then during training concatenated with source images to form an input with two channels (denoted G+D). The result is significantly worse than the baseline.

\begin{table}[htb]
\centering
\caption[AMOS Patches mAP scores of HardNet trained on data of different modality]{AMOS Patches mAP scores of HardNet trained on data of different modality. G denotes grayscale, D denotes (estimanted mono-)depth.}

\begin{tabular}{clc}
\toprule
Modality & Model & mAP\\
\midrule
G & HardNet8 & 49.09 \\
G+D & HardNet8, input channels = 2 & 45.74 \\
RGB & HardNet8, input channels = 3 & 47.56 \\
RGB & HardNet8, learned color conversion & \textbf{49.31} \\
\bottomrule
\end{tabular}
\label{tab:modality}
   
\end{table}

\paragraph{Number of AMOS Views}

\begin{figure}[htb]
\centering

\includegraphics[width=0.4\linewidth]{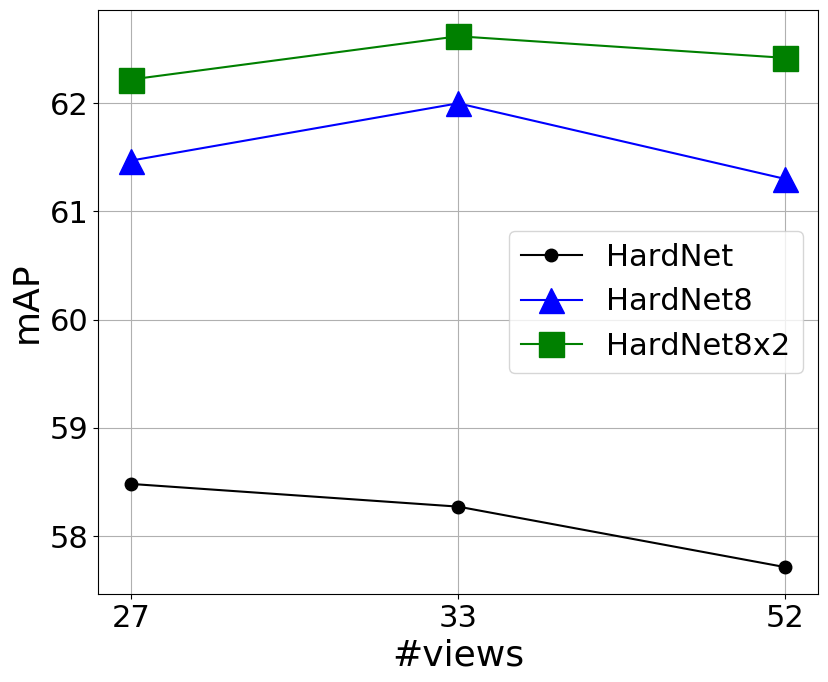}

\caption[Evaluation of HardNet trained on subsets of AMOS Views.]{Evaluation of HardNet trained on subsets of AMOS Views. We measure mean average precision (mAP) on the HPatches benchmark.}
% \caption[Evaluating AMOS Patches datasets]{AMOS Views variants. mAP measure is evaluated on HPatches if we train HardNet architecture on increasingly larger dataset.}
\label{fig:camsubsets}
\end{figure}

During manual pruning in the AMOS Patches extraction pipeline we have the possibility to select up to 146 views. To check if more views are beneficial or not, we prepared 3 versions which are ordered with respect to subset relation. In this experiment we train for 20 epochs with batch size = 3072 and compare three architectures. One can see in Figure \ref{fig:camsubsets} that using a bigger training dataset does not necessarily improve performance. We make a similar observation also later in the case of IMW PhotoTourism data. Based on this result, we recommend the smaller subset for training of HardNet and the medium dataset for training of HardNet8.

\begin{figure}[htb]
\centering

\begin{subfigure}[b]{0.475\textwidth}
  \frame{\includegraphics[width=\linewidth]{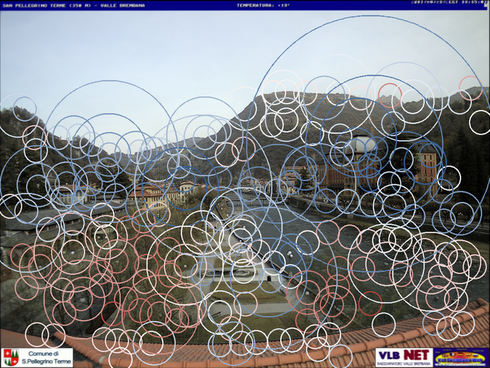}}
%   \caption{}
\end{subfigure}
\hfill
\begin{subfigure}[b]{0.475\textwidth}
  \frame{\includegraphics[width=\linewidth]{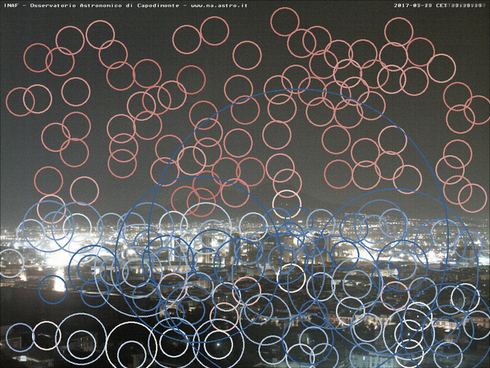}}
%   \caption{}
\end{subfigure}
\vskip\baselineskip
\begin{subfigure}[b]{0.475\textwidth}
  \frame{\includegraphics[width=\linewidth]{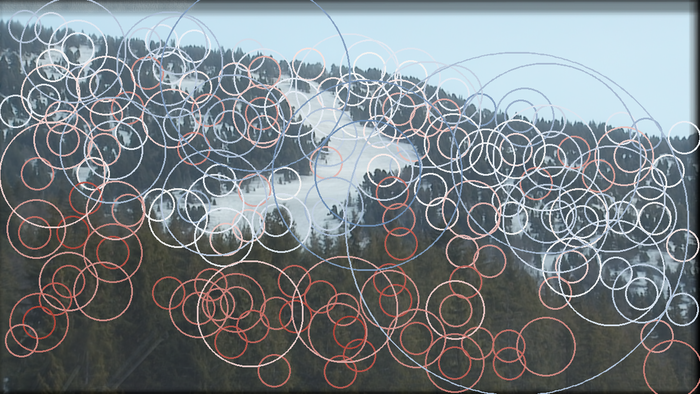}}
%   \caption{}
\end{subfigure}
\hfill
\begin{subfigure}[b]{0.475\textwidth}
  \frame{\includegraphics[width=\linewidth]{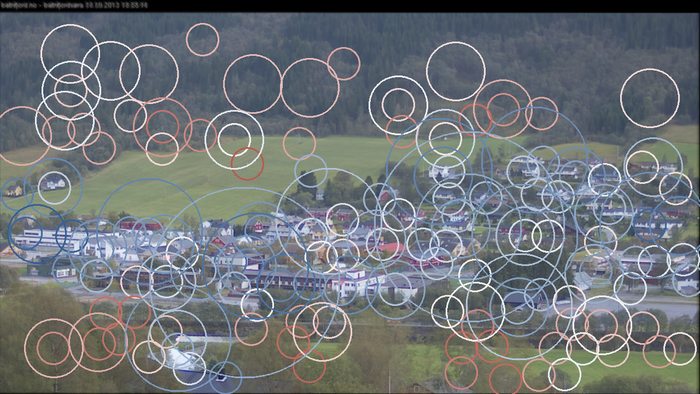}}
%   \caption{}
\end{subfigure}

\caption[Visualization of the loss function values on AMOS Views images at the end of training.]{Visualization of the loss function values on AMOS Views images at the end of training. Values range from 0.8 (blue), over 1 (white) to 1.2 (red). Note that value above 1 means that the negative example is closer than positive, i.e. wrong match. Here margin in the loss function \cite{mishchuk2017working} is equal to 1.}
\label{fig:loss}
\end{figure}

\paragraph{Distribution of Loss Values}
We noticed that during training the average values of the loss function per patch set are distributed according to where the patch was extracted, see Figure \ref{fig:loss}. It holds approximately that patch sets with higher average loss are in parts of the image which view dynamic content such as trees, grass or sky. And vice versa, patch sets with low average loss are positioned on static content such as buildings or walkways.

\subsection{Experiments with IMW Phototourism}
\label{sec:expsimw}
\paragraph{Dataset Reduction}
As we mentioned earlier, some of the IMW PT datasets are very large. The extracted patches from Colosseum Exterior scene occupy 32.6 GB, while patches from Trevi Fountain scene -- even 65.4 GB. The goal of this experiment is to explore a data reduction technique and compare the smaller versions to the full dataset. Note that we made a related observation with AMOS Patches, where we found that using smaller number of cameras might be beneficial.

The description of the dataset reduction procedure follows. In the first step we sample batches randomly in the same manner as during training, obtain the descriptors for each batch by using HardNet pretrained on AMOS Patches, and save the observed differences $$e(p_1)=d(p_1,p_2)- min(d_{\textrm{neg1}}, d_{\textrm{neg2}})$$ for each patch $p_1$. Let $\bar{e}_i$ be the mean value of such differences for patch set $i$ over all patches it contains and $$\bar{e}=\frac{1}{N}\sum\limits_{j=1}^{N}\bar{e}_i$$ be the mean value over all $N$ patch sets in the dataset. Then we sort the patch sets $i=1,\dots,N$ in ascending order according to the value of $l_i=|\bar{e}_i - \bar{e}|$.
To reduce the dataset to $X$ number of patches, we add only those patch sets, which have the lowest distance $l_i$, so that the overall number of patches is less or equal to $X$.  Such patch sets are of "medium" hardness relative to the given dataset. The remaining patches are then added randomly from the first unused patch set with the lowest $l_i$.

We also tried other sorting approaches to reducing a dataset, which we show in Table \ref{tab:reduction}. "Medium" refers to the above described procedure. In "Low" and "High" we select patch sets with the lowest (easy patches) and highest (hard patches) $\bar{e}_i$ values respectively. Note that taking the hardest patches leads to the worst performance on all datasets.

% "Lower" prioritizes patch sets with lower $\bar{e}_i$ values, "Upper" denotes the opposite case. "Medium" leads to better results on IMW PT and "Lower" to better results on HPatches and AMOS Patches.

\begin{table}[htb]
\centering
\caption[Evaluation of dataset reduction approaches.]{Evaluation of dataset reduction approaches. The first column denotes which patch sets are selected first based on the $\bar{e}_i$ values. Each reduced version contains 1.6 million patches. Mean average precision (mAP) is measured on HPatches and AMOS Patches. In the IMW PT benchmark we measure mean average accuracy (mAA) with the threshold of 10\degree.}
% \caption[Approaches to reducing a dataset: ]{Approaches to reducing a dataset. The first column denotes which patch sets are selected first based on the $\bar{e}_i$ values. Each reduced version contains 1.6 million patches. Mean average precision (mAP) is measured on HPatches and AMOS Patches. In the IMW PT benchmark we measure mean average accuracy (mAA) with the threshold of 10\degree.}

\begin{tabular}{lccc}
\toprule
$\bar{e}_i$ & HPatches & AMOS Patches & IMW PT \\
\midrule

Low & \textbf{43.20}	&\textbf{47.05}&	68.78 \\
Medium & 42.38	&	45.91&	\textbf{69.77} \\
High & 40.64&	42.99	&	66.77 \\
\bottomrule

\end{tabular}
\label{tab:reduction}
\end{table}

\begin{figure}[htb]
\centering

\begin{subfigure}[b]{0.49\linewidth}
\centering
  \includegraphics[width=0.98\linewidth]{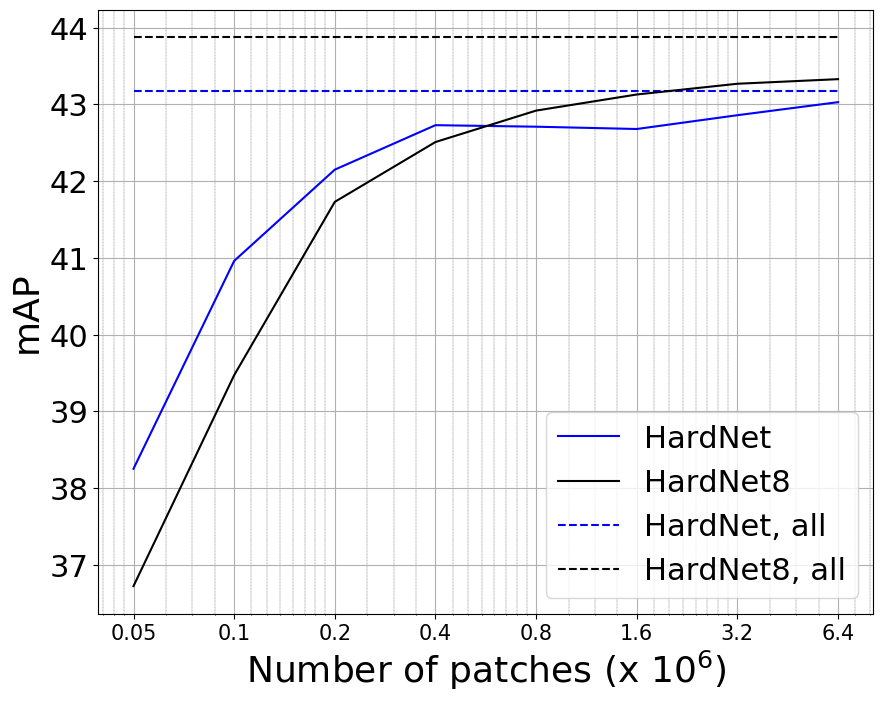}
  \caption{HPatches full split.}
\end{subfigure}
\begin{subfigure}[b]{0.49\linewidth}
\centering
  \includegraphics[width=0.98\linewidth]{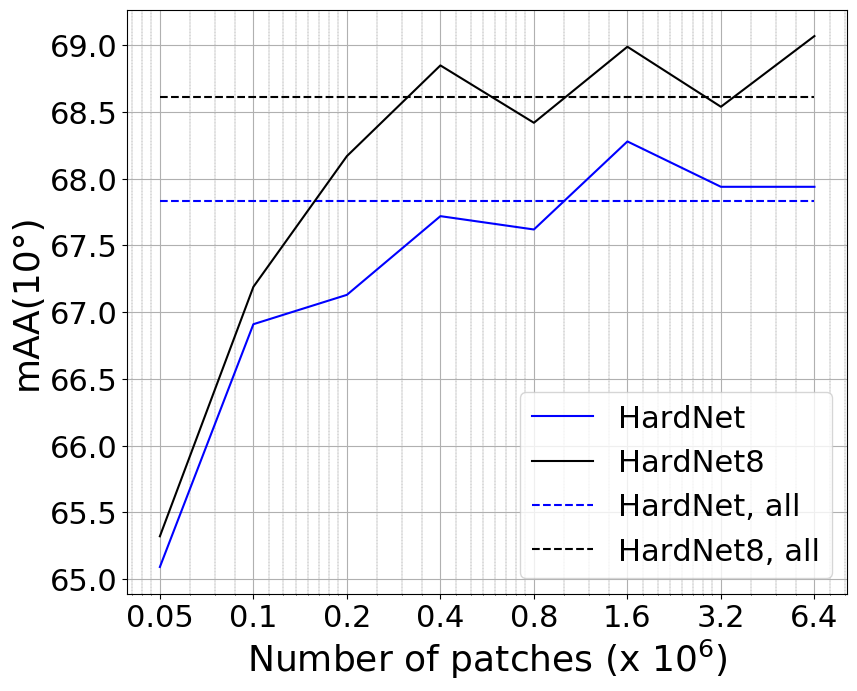}
  \caption{IMW PT, validation set.}
\end{subfigure}

\caption[Evaluation of the dataset reduction method.]{Evaluation of the dataset reduction method. The Trevi dataset is used. Dashed lines represent performance of the model trained on full dataset. The measure is mean average precision (mAP) on HPatches and mean average accuracy (mAA) with the threshold of 10\degree on IMW PT.}
% \caption{Reducing size of the Trevi dataset from IMW PT. Dashed lines represent performance of the model trained on full dataset.}
\label{fig:reduction}
\end{figure}

Figure~\ref{fig:reduction}  shows that on the IMW PT benchmark some of the reduced versions lead to better performance. Bigger architecture achieves better score with any number of patches. On the contrary, on HPatches the result is consistently worse on all reduced versions than if we train on the full dataset. For smaller number of patches the smaller architecture gives better mAP measure, while for more patches it becomes the opposite.

\paragraph{Comparison of Scenes}
Here we make use of the described procedure of dataset reduction and of the observation that training on a smaller version leads to comparable, and sometimes better, performance. We consider several datasets from the collection of IMW PT benchmark, for each make a smaller version which contains exactly 1.6 million patches and train the HardNet architecture for 10 epochs, 5 millions samples each. The results are listed in tab \ref{tab:datasets}. One can notice that good performance on one benchmark does not imply similar result on the other two.

\begin{table}[htb]
\centering
\caption[Training datasets comparison.]{Training datasets comparison. HardNet is trained on a dataset specified in the left column. Each dataset contains exactly 1.6 million patches. Mean average precision (mAP) is measured on HPatches and AMOS Patches. On the IMW PT benchmark we measure mean average accuracy (mAA) with the threshold of 10\degree.}
   
\begin{tabular}{lccc}
\toprule
Dataset & HPatches & IMW PT & AMOS Patches \\
\midrule
Buckingham Palace & 35.99 & 67.87 & 43.42 \\
Colosseum Exterior & 42.38 & \textbf{69.77} & 45.91 \\
Brandenburg Gate & 40.58 & 68.54 & 44.12 \\
Hagia Sophia Interior & \textbf{49.27} & 67.46 & 39.86 \\
Trevi Fountain & 42.68 & 68.28 & \textbf{46.78} \\
Prague Old Town Square & 42.63 & 67.42 & 46.46 \\
\bottomrule

\end{tabular}
\label{tab:datasets}
\end{table}

\subsection{Combining Datasets}
When multiple diverse datasets are available, a logical next step is to try to combine them for training. For this purpose we developed and used own software\footnote{Public version available at https://github.com/pultarmi/HardNet\_MultiDataset} and summarize our findings in Table~\ref{tab:combining} for Liberty, Notredame and AMOS Patches datasets. We compare three methods: In "Per Epoch" we split each epoch into $D$ contiguous parts -- $D$ is number of datasets -- and in each part we sample always from a single dataset. In "Per Batch" each batch is composed of patches from a single random dataset. In "In Batch" we split each batch into $D$ blocks, where each block is composed of patches from a different dataset. "Per Batch" method most often consistently achieves better results on all benchmarks, while "Per Epoch" tends to forget the dataset, which comes first in the last epoch, as can be seen in Table~\ref{tab:combining} in the case of Lib+AMOS.

Combining the datasets leads to some improvement. However, one has to be careful about several things. First, in some situations we often experienced the opposite behaviour -- e.g. HardNet8 trained on Colosseum achieves 70.75 mAA(10$\degree$), while adding Liberty leads to 70.21 mAA(10$\degree$). This negative effect can, in a sense, be reverted using compression of descriptors (explained in section \ref{sec:compression}), as we observed that dimensionality reduction works better if the model was trained on more datasets. Also, if AMOS Patches is present in the ensemble of datasets, it tends to influence the result significantly by improving result in HPatches and AMOS Patches, while worsening it in IMW PT. For this reason, we advise to tune the probability of sampling from AMOS using e.g. settings available in our software.

\begin{table}[htb]
\centering
\caption[Combining datasets: Influence of the training dataset(s) and combining methods on the measured metrics.]{Combining datasets: Influence of the training dataset(s) and combining methods on the measured metrics. The HardNet architecture is used. Lib+AMOS denotes that Liberty is sampled first in each epoch, for AMOS+Lib it is the opposite. Mean average precision (mAP) is measured on HPatches and AMOS Patches. In the IMW PT benchmark we measure mean average accuracy (mAA) with the threshold of 10\degree.}
%HardNet architecture trained on a single or a combination of datasets. 

\begin{tabular}{lcccc}
\toprule
Train on & Combine & HPatches & AMOS Patches & IMW PT \\
\midrule

Liberty &--& 52.96 & 44.21 & 68.17 \\
Notredame &--& 52.44 & 44.65 & 68.56 \\
AMOS &--& 58.27 & 46.96 & 61.50 \\
\midrule
Lib+Notre & Per Epoch & 53.14 & 45.63 & 68.82 \\
 & Per Batch & \textbf{53.92} & \textbf{45.84} & \textbf{68.89} \\
 & In Batch & 53.02 & 45.02 & 68.14 \\
\midrule
Lib+AMOS & Per Epoch & 59.23&	48.62 & 65.44  \\
\midrule
AMOS+Lib & Per Epoch & 56.63 & 46.66 & \textbf{68.25}  \\
 & Per Batch & \textbf{59.28} & \textbf{48.43} & 66.47 \\
 & In Batch & 57.77 & 46.77 & 66.04 \\
\bottomrule

\end{tabular}
\label{tab:combining}
\end{table}

\section{Architecture}
\label{sec:architectures}
The architecture of a model is one of the paramount factors influencing the performance of the descriptor. A lot of work in the field of computer vision is focused on discovering architectures that achieve higher performance, are faster or more compact. However, there is a paucity of literature available describing new architectures for a local feature descriptor. A convolutional architecture was introduced in \cite{l2net} and was adopted by others \cite{mishchuk2017working, geodesc2018}. There has been little effort to find a better architecture since. Better architecture can generally be obtained either by manual search or by using an algorithmic solution. Here we explore both approaches.

\subsection{Automated Search}
Differentiable Architecture Search (DARTS)~\cite{liu2018darts} is an algorithm which searches a space of architectures formulated in a continuous manner by using gradient descent. The bilevel optimization problem it approximates can be expressed as follows. Credit:~\cite{liu2018darts}:

\begin{align}
&\min\limits_{\alpha} \mathcal{L}_{val}(w^*(\alpha), \alpha),\\
&\textrm{s.t.}\ w^*(\alpha)= \arg \min\limits_w \mathcal{L}_{train}(w, \alpha), \nonumber
\end{align}

where $\mathcal{L}_{train}$ and $\mathcal{L}_{val}$ is the loss function on the training and validation set, $\alpha$ is the architecture and $w$ are the associated weights. The search space is represented by a sequence of cells, which are composed of $k$ ordered nodes. The input to each node (called a connection or edge) is equal to the sum of its two predecessors. The output of a cell is defined as a concatenation of all its nodes. The edges between the nodes correspond to operations that are to be learned. During architecture search each of these are assigned parameters that are weighted using the softmax operator.

After the search procedure is finished, the operation corresponding to the highest weight is assigned to each edge. It is recommended to use a larger number of cells at this point due to smaller resource requirements.

In our experiments we make use of the publicly available implementation\footnote{Available at https://github.com/quark0/darts} and plug in our data loaders, so that Liberty and Notredame from UBC Phototour~\cite{snavely2008modeling, goesele2007multi} is the training and validation dataset respectively. We run architecture search for 40 epochs, 100 000 samples each, learning rate = 2.5, batch size = 64. Other settings are kept default.

\begin{figure}[htb]
\centering

\begin{subfigure}[b]{0.49\linewidth}
\centering
  \includegraphics[width=0.98\linewidth]{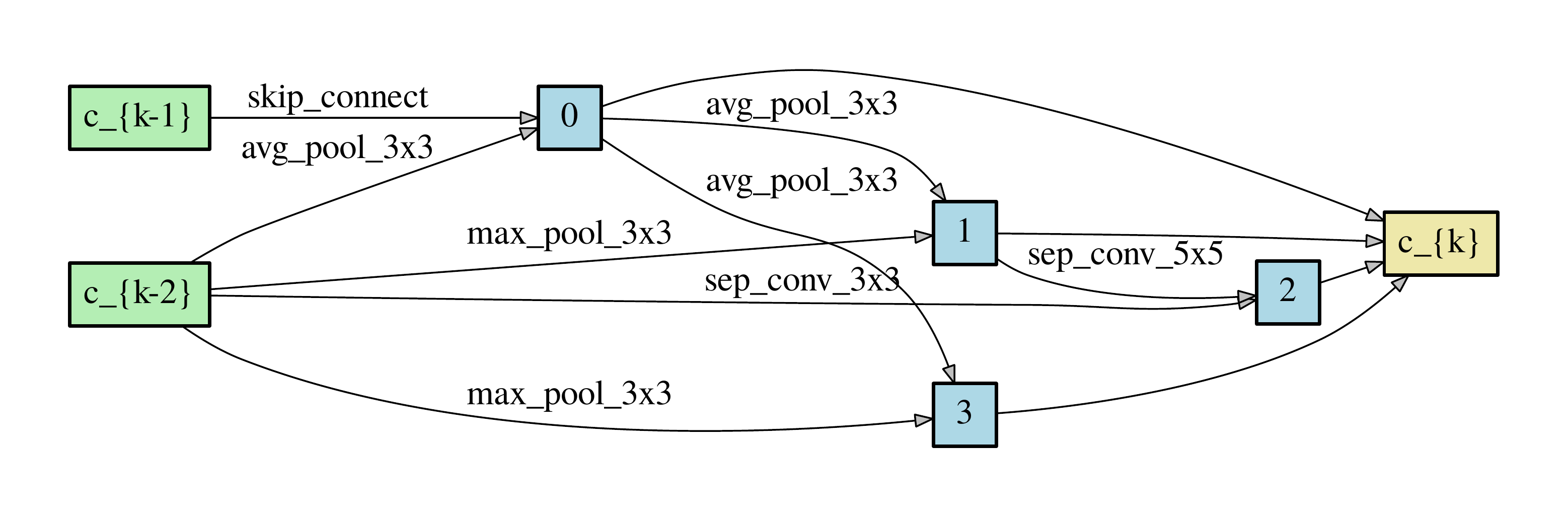}
  \caption{Reduction module.}
\end{subfigure}
\begin{subfigure}[b]{0.49\linewidth}
\centering
  \includegraphics[width=0.98\linewidth]{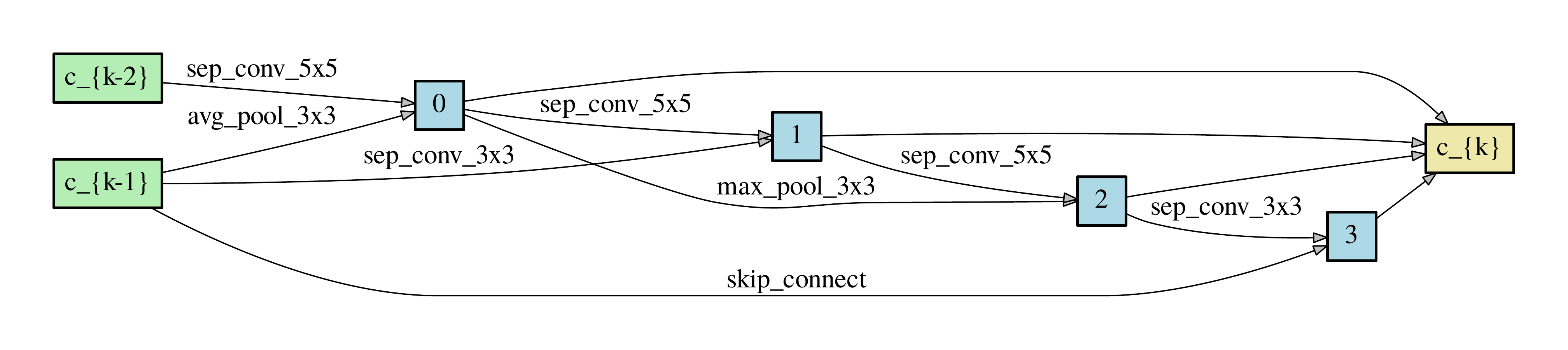}
  \caption{Normal module.}
\end{subfigure}
\hfill\par

\caption[Modules found by the DARTS algorithm.]{Modules found by the DARTS algorithm visualized using the provided tool~\cite{liu2018darts}.}
% \caption[Modules found by the DARTS algorithm.]{Modules constituting the architecture obtained by DARTS algorithm visualized using the provided tool~\cite{liu2018darts}.}
\label{fig:darts}
\end{figure}

Subsequent training is performed for 40 epochs, 500 000 tuples each, with batch size = 128 and learning rate = 0.025. In Table \ref{tab:dartstab} we list the results for several runs and compare with HardNet. In Figure \ref{fig:darts} we present the found cells constituting the architecture DARTS 1. Reduction module is used in two positions in the architecture (1/3 and 2/3 of the length), all other cells are normal. The results of the DARTS models are significantly worse than those of HardNet. It might be for example due to small batch size - higher is not feasible due to GPU memory. Also, the method could be improved by changing the search space. We leave such modifications for future work.

\begin{table}[htb]
\centering
\caption[Evaluation of models found by the DARTS algorithm.]{Evaluation of models found by the DARTS algorithm, one model per row. Several trainings were run due to the recommendation in \cite{liu2018darts}. We measure mean average precision (mAP) for HPatches and AMOS Patches and mean average accuracy mAA(10$\degree$) for the CVPR IMW benchmark.}

\begin{tabular}{lccc}
\toprule
Architecture & HPatches & AMOS Patches & IMW PT \\
\midrule

DARTS 1 & 36.56 & 26.42 & 57.38 \\
DARTS 2 & 28.48 & 17.41 & 46.53 \\
DARTS 3 & 24.54 & 13.36 & 40.76 \\
\midrule
HardNet & \textbf{52.96} & \textbf{44.21} & \textbf{68.17} \\
\bottomrule

\end{tabular}
\label{tab:dartstab}
\end{table}

\subsection{Manual Search}
%the set of considered operations on the edges

\paragraph{VGG~\cite{simonyan2014very} Style}
HardNet is a VGG style network with blocks of convolutional layers of decreasing spatial size and an increasing number of channels. Here we introduce and evaluate several modifications to the architecture. With respect to the original version, HardNet7x2 contains two times more channels in each layer, HardNet8 has one convolutional layer added in the third block and HardNet9 contains two more layers, one in the second and one the third block. HardNet8x2 has two times more channels in each layer than HardNet8. See tab \ref{fig:archhard} for a detailed overview.

In Table \ref{tab:manualhard} we list the results for each architecture

\newcommand\cols{5}

\begin{table*}[htb]
\centering
\caption[Variants of the HardNet architecture]{Variants of the HardNet architecture. Each convolutional layer is followed by batch normalization and ReLU activation.}
\footnotesize
\begin{tabular}{ccccc}

\toprule
 HardNet & HardNet7x2 & HardNet8 & HardNet8x2 & HardNet9 \\
\midrule
 1.3M param & 5.3M param & 4.7M param & 19M param & 5.3 param \\
\midrule
\multicolumn{\cols}{c}{Block 1, spatial size 32x32}\\
\midrule
Conv 3x3x32 & Conv 3x3x64 & Conv 3x3x32 & Conv 3x3x64 & Conv 3x3x32 \\
Conv 3x3x32 & Conv 3x3x64 & Conv 3x3x32 & Conv 3x3x64 & Conv 3x3x32 \\
\midrule
\multicolumn{\cols}{c}{Block 2, spatial size 16x16}\\
\midrule
Conv 3x3x64/2 & Conv 3x3x128/2 & Conv 3x3x64/2 & Conv 3x3x128/2 & Conv 3x3x64/2 \\
Conv 3x3x64 & Conv 3x3x128 & Conv 3x3x64 & Conv 3x3x128 & Conv 3x3x64 \\
- & - & - & - & Conv 3x3x128 \\
%\cmidrule(r){1-4}
\midrule
\multicolumn{\cols}{c}{Block 3, spatial size 8x8}\\
\midrule
Conv 3x3x128/2 & Conv 3x3x256/2 & Conv 3x3x128/2 & Conv 3x3x256/2 & Conv 3x3x128/2 \\
Conv 3x3x128 & Conv 3x3x256 & Conv 3x3x128 & Conv 3x3x256 & Conv 3x3x256 \\
- & - & Conv 3x3x256 & Conv 3x3x512 & Conv 3x3x256\\
%\cmidrule(r){1-4}
\midrule
\multicolumn{\cols}{c}{Global pooling block}\\
\midrule
\multicolumn{\cols}{c}{Dropout(0.3)}\\
\midrule
Conv 8x8x128 & Conv 8x8x256 & Conv 8x8x256 & Conv 8x8x256 & Conv 8x8x256 \\
\midrule
\multicolumn{\cols}{c}{Flatten(), L2Norm()}\\
\bottomrule

\end{tabular}

\label{fig:archhard}
\end{table*}

\begin{table}[htb]
\centering
\caption[Evaluation of the variants of the HardNet architecture]{Evaluation of the variants of the HardNet architecture. Mean average precision (mAP) is measured on HPatches and AMOS Patches. In the CVPR IMW benchmark we measure mean average accuracy (mAA) with the threshold of 10\degree}

\begin{tabular}{lccc}
\toprule
Architecture & HPatches & AMOS Patches & IMW PT \\
\midrule

HardNet & 52.96 & 44.21 & 68.17 \\
HardNet7x2 & \textbf{54.56} & \textbf{44.39} & 68.67 \\
HardNet8 & 53.67 & 43.77 & \textbf{69.43} \\
HardNet8x2 & 54.55 & 42.96 & 69.18 \\
HardNet9 & 54.21 & 42.85 & 69.34 \\
\bottomrule

\end{tabular}
\label{tab:manualhard}
\end{table}

\paragraph{ResNet~\cite{he2016deep} Style}
Another popular architecture type is ResNet. It consists of blocks of convolutional layers called ResBlock. The output from each block is summed with its input, so the network learns an incremental change instead of a direct transformation. In Table \ref{fig:arch_res} we list the architectures we tested. Evaluation of these models is shown in Table \ref{tab:evalres}. Each model was trained on the Liberty dataset for 20 epochs, 5 million samples each. The results are almost on par with those of HardNet models. However, in none of our experiments it was superior and it has a higher GPU memory requirement, so we can not advise to use this architecture type for the learning of a local feature descriptor.

\renewcommand\cols{3}

\begin{table*}[htb]
\centering
\caption[ResNet architectures]{ResNet architectures. Each convolutional layer is followed by batch normalization and ReLU activation.}
\footnotesize
\begin{tabular}{ccc}

\toprule
ResNet2 & ResNet3 & ResNet3x \\
\midrule
1.7 param & 2M params & 2.1M params \\
\midrule
\multicolumn{\cols}{c}{Block 1, spatial size 32x32}\\
\midrule
Conv 3x3x32 & Conv 3x3x16 & Conv 3x3x32 \\
Conv 3x3x64 & Conv 3x3x32 & Conv 3x3x64 \\
\midrule
\multicolumn{\cols}{c}{Block 2, spatial size 16x16}\\
\midrule
ResBlock2/2 64 & ResBlock3/2 64 & ResBlock3/2 64 \\
ResBlock2 64 & ResBlock3 64 & ResBlock3 64 \\
%\cmidrule(r){1-4}
\midrule
\multicolumn{\cols}{c}{Block 3, spatial size 8x8}\\
\midrule
ResBlock2/2 128 & ResBlock3/2 128 & ResBlock3/2 128 \\
ResBlock2 128 & ResBlock3 128 & ResBlock3 128 \\
%\cmidrule(r){1-4}
\midrule
\multicolumn{\cols}{c}{Global pooling block}\\
\midrule
\multicolumn{\cols}{c}{Dropout(0.3)}\\
\midrule
Conv 8x8x128 & Conv 8x8x128 & Conv 8x8x128 \\
\midrule
\multicolumn{\cols}{c}{Flatten(), L2Norm()}\\
\bottomrule

\end{tabular}

\label{fig:arch_res}
\end{table*}

\begin{figure}[htb]
\centering

\begin{subfigure}[b]{0.22\linewidth}
\centering
  \includegraphics[width=0.7\linewidth]{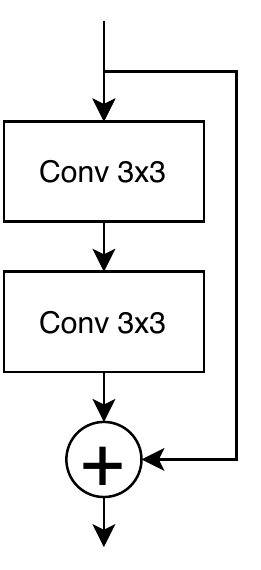}
  \caption{ResBlock2}
\end{subfigure}
\begin{subfigure}[b]{0.22\linewidth}
\centering
  \includegraphics[width=0.7\linewidth]{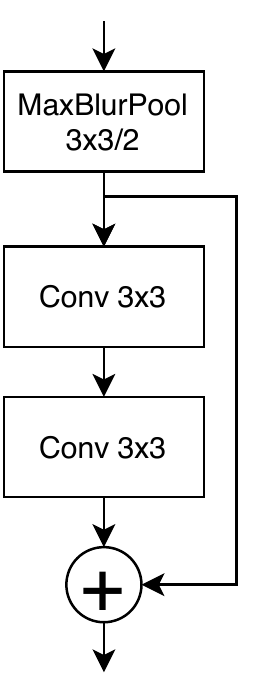}
  \caption{ResBlock2/2}
\end{subfigure}
\begin{subfigure}[b]{0.22\linewidth}
\centering
  \includegraphics[width=0.7\linewidth]{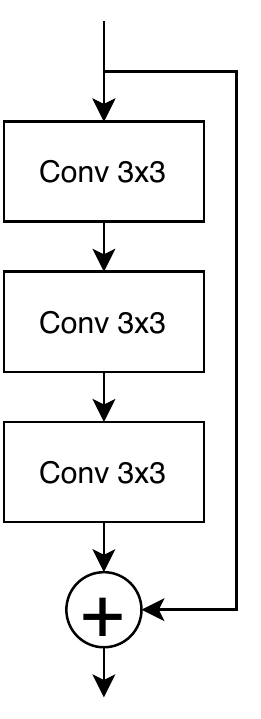}
  \caption{ResBlock3}
\end{subfigure}
\begin{subfigure}[b]{0.22\linewidth}
\centering
  \includegraphics[width=0.7\linewidth]{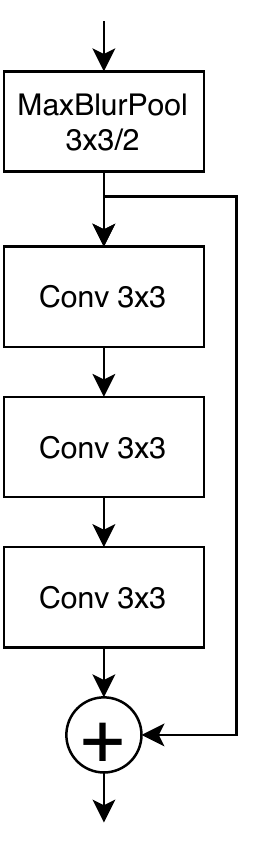}
  \caption{ResBlock3/2}
\end{subfigure}

\hfill\par

\caption{ResNet modules.}
% \caption[Modules used in our implementation of ResNet.]{Modules used in our implementation of ResNet.}
\label{fig:resblock}
\end{figure}

\begin{table}[htb]
\centering
\caption[Evaluation of the ResNet architectures.]{Evaluation of the ResNet architectures together with the baseline HardNet. All models are trained on Liberty. Mean average precision (mAP) is measured on HPatches and AMOS Patches. On the CVPR IMW benchmark we measure mean average accuracy (mAA) with the threshold = 10\degree.}
   
\begin{tabular}{lccc}
\toprule
Architecture & HPatches & AMOS Patches & IMW PT \\
\midrule

ResNet2 & 52.63 & 40.57 & 68.31 \\
ResNet3 & 52.96 & 40.72 & 69.01 \\
ResNet3x & 52.25 & 38.50 & 68.27 \\
\midrule
HardNet8 & \textbf{53.67} & \textbf{43.77} & \textbf{69.43} \\
\bottomrule

\end{tabular}
\label{tab:evalres}
\end{table}

\paragraph{Input Size}
The input to the HardNet architecture is of size 32$\times$32 pixels. Due to the observation that enlarging the model architecture further from HardNet8 does not lead to better performance, we perform the following experiment. We make minor adjustments to the HardNet8 architecture so that it expects inputs of a bigger size. First, we change the stride to 2 in the penultimate layer, then input size is 64$\times$64 pixels. If we also set the kernel size of the last convolutional layer to 6$\times$6 pixels, the input size is then 48$\times$48 pixels. We compare these three models in Table \ref{tab:isize}. Notice the two opposite trends: bigger input size leads to better performance on HPatches and AMOS Patches, but such a model gives inferior results on the IMW PT benchmark.

\begin{table}[htb]
\centering
\caption[Evaluation of HardNet variants with different input size.]{Evaluation of HardNet variants with different input size. Mean average precision (mAP) is measured on HPatches and AMOS Patches. In the IMW PT benchmark we measure mean average accuracy (mAA) with the threshold of 10\degree.}
% \caption[Evaluation of HardNet variants with different input size.]{Evaluation of HardNet variants with different input size. Mean average precision (mAP) and mean average accuracy (mAA) for variants of the HardNet architecture}
% \caption[Changing the input size.]{Changing the input size. Mean average precision (mAP) and mean average accuracy (mAA) for variants of the HardNet architecture}

\begin{tabular}{lccc}
\toprule
Model & HPatches & AMOS Patches & IMW PT \\
\midrule

HardNet8 & 52.37 & 42.42 & \textbf{69.06} \\
HardNet8, input size = 48 & 53.34 & 43.12 & 68.41 \\
HardNet8, input size = 64 & \textbf{54.19} & \textbf{43.98} & 68.25\\
\bottomrule

\end{tabular}
\label{tab:isize}
\end{table}

\subsection{Final Pooling}
The HardNet architecture uses global pooling implemented by a 8x8x128 convolution as the last layer in the model -- i.e. in fact a fully connected linear layer -- unlike SIFT, which uses local pooling, where the output vector is composed of 16 blocks based on the relative position in the input patch. In this section we compare different architectures and try several pooling approaches. All models are trained on the Liberty dataset for 20 epochs, 5 million samples each.

%Note that while SIFT uses local pooling, i.e. the output vector is composed of 16 blocks based on the relative position in the input patch, the HardNet architecture uses global pooling implemented by a 8x8x128 convolution as the last layer in the model -- i.e. in fact a fully connected linear layer. In this section we compare different architectures and try several pooling approaches. All models are trained on the Liberty dataset for 20 epochs, 5 million samples each.

\paragraph{Local pooling}
In this experiment we try to mimic the mechanism of SIFT and replace the last convolutional layer in order to increase its stride so that the convolution is performed over separate blocks. In Table \ref{tab:finalpool} we can observe that the performance of these models is worse than of the baseline.
\paragraph{Increased Receptive Field}
We also test modifications to the HardNet8 architecture which increase its receptive field. In Table \ref{tab:finalpool} we can observe that such architectures do not outperform the baseline. Enlarging the architecture by 3 layers even fails to learn.

\begin{table}[htb]
\centering
\caption[Evaluation of final pooling variants.]{Final pooling variants. We list modifications to the HardNet8 architecture by substituting the last convolutional layer. The last row represents no modification. X$^y$ means the layer X is stacked $y$ times. We write n/c for models which failed to learn. Mean average precision (mAP) is measured on HPatches and AMOS Patches. On the IMW PT benchmark we measure mean average accuracy (mAA) with the threshold of 10\degree.}
   
\begin{tabular}{lccc}
\toprule
Final Layer(s) & HPatches & AMOS Patches & IMW PT \\
\midrule
\multicolumn{4}{c}{Local Pooling}\\
\midrule
Conv 4x4x16/4 & 53.28 & 44.50 & 68.50 \\
Conv 3x3x16/2, p=1 & 52.76 & \textbf{44.78} & 68.36 \\
\midrule
\multicolumn{4}{c}{Increased Receptive Field}\\
\midrule
(Conv 3x3/2, p=1)$^1$ $\rightarrow$ Conv 4x4, p=0  & 52.31 & 42.00 & 68.29 \\
(Conv 3x3/2, p=1)$^2$ $\rightarrow$ Conv 2x2, p=0 & 50.25 & 39.14 & 67.62 \\
(Conv 3x3/1, p=0)$^3$ $\rightarrow$ Conv 2x2, p=0 & n/c & n/c & n/c \\
\midrule
\multicolumn{4}{c}{Non-learned}\\
\midrule
MaxPool 8x8 & 44.23 & 38.58& 67.11 \\
AvgPool 8x8  & n/c & n/c & n/c \\
\midrule
\multicolumn{4}{c}{Global Pooling (original)}\\
\midrule
Conv 8x8x256 & \textbf{53.67} & 43.77 & \textbf{69.43} \\
\bottomrule

\end{tabular}
\label{tab:finalpool}
\end{table}

\paragraph{Non-learned Pooling}
Another possibility is to compute the maximum or average per input channel. MaxPool achieves worse result and AvgPool fails to learn, see Table~\ref{tab:finalpool}.

\section{Compression of Embeddings}
\label{sec:compression}
\begin{figure}[htb]
\centering

\begin{subfigure}[b]{0.49\linewidth}
\centering
  \includegraphics[width=0.98\linewidth]{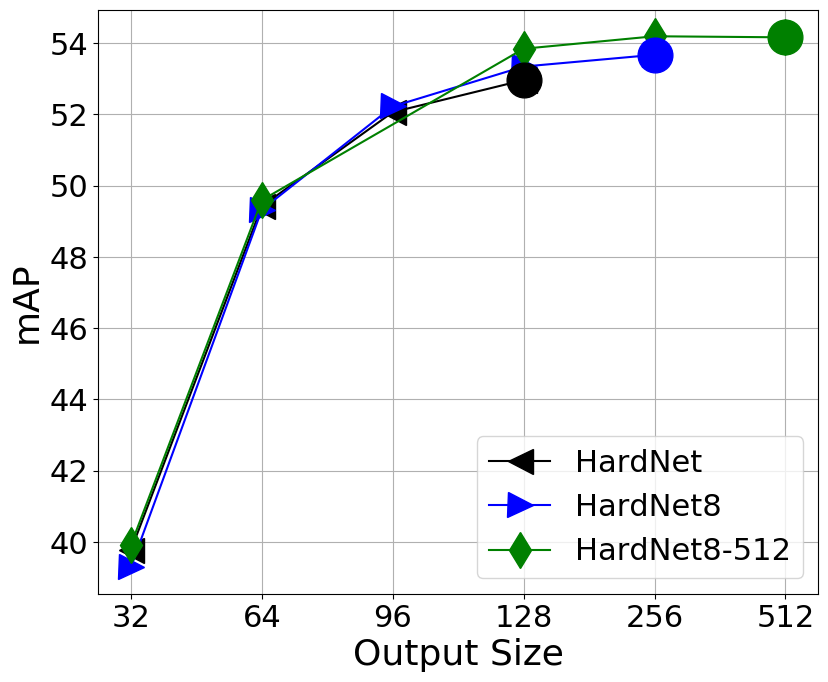}
  \caption{HPatches full split.}
\end{subfigure}
\begin{subfigure}[b]{0.49\linewidth}
\centering
  \includegraphics[width=0.98\linewidth]{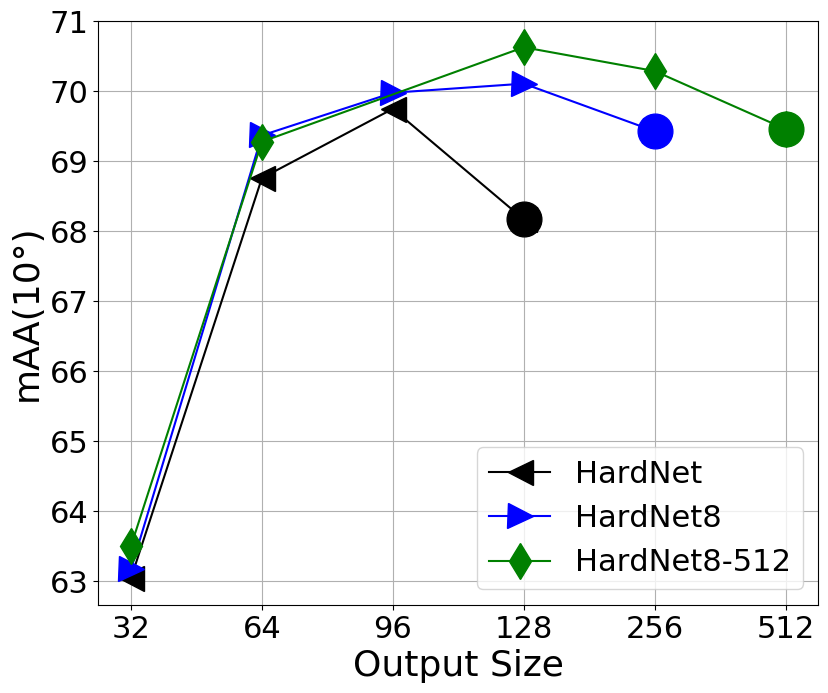}
  \caption{IMW PT, validation set.}
\end{subfigure}

\caption[Impact of the embeddings compression.]{Impact of the embeddings compression. PCA dimensionality reduction method is used. HardNet8-512 denotes HardNet8 with the number of channels in the last layer changed to 512. The measure is mean average precision (mAP) on HPatches and mean average accuracy (mAA) with the threshold of 10\degree on IMW PT.
Original, non-compressed dimension is marked with a circle.}
% \caption[Compression of embeddings: Infuence of the output size on the measured metrics.]{Compression of embeddings: Infuence of the output size on the measured metrics. PCA dimensionality reduction method is used. HardNet8-512 denotes HardNet8 with the number of channels in the last layer changed to 512. The measure is mean average precision (mAP) on HPatches and mean average accuracy (mAA) on IMW PT.}
\label{fig:pca}
\end{figure}

During manual architecture search we found that some of the larger models, e.g. HardNet8, achieve better performance. These architectures however also output longer output vector, which brings disadvantages such as higher memory usage and slower nearest-neighbour search. If we reduce the output size of HardNet8 to 128 to match the vector length of SIFT, we get inferior results - 68.75 mAA(10\degree) vs 69.43 mAA(10\degree) on the IMW PT benchmark. Another way to decrease the output size is to use a  dimensionality reduction technique. Here we make use of principal component analysis (PCA) to compress the feature embeddings. See in Figure \ref{fig:pca} that dimensionality reduction is beneficial. We can observe that the best combination is HardNet8 with output size 512 followed by compression to size 128. In this experiment we train on the Liberty dataset from UBC Phototour and we fit PCA on the model outputs from the training data.

\section{Loss Function}
\label{sec:loss}
In this section we explore several modifications to the hard-in-batch triplet margin loss function. First, we find the optimal value for the margin in the original version, viz eq. \ref{eq:lossfc}. Then we introduce a generalization of the loss by sampling n-tuples of positives instead of pairs. Finally, we test an existing implementation of a variant with a dynamic margin.

\begin{figure}[htb]
\centering

\begin{subfigure}[b]{0.49\linewidth}
\centering
  \includegraphics[width=0.98\linewidth]{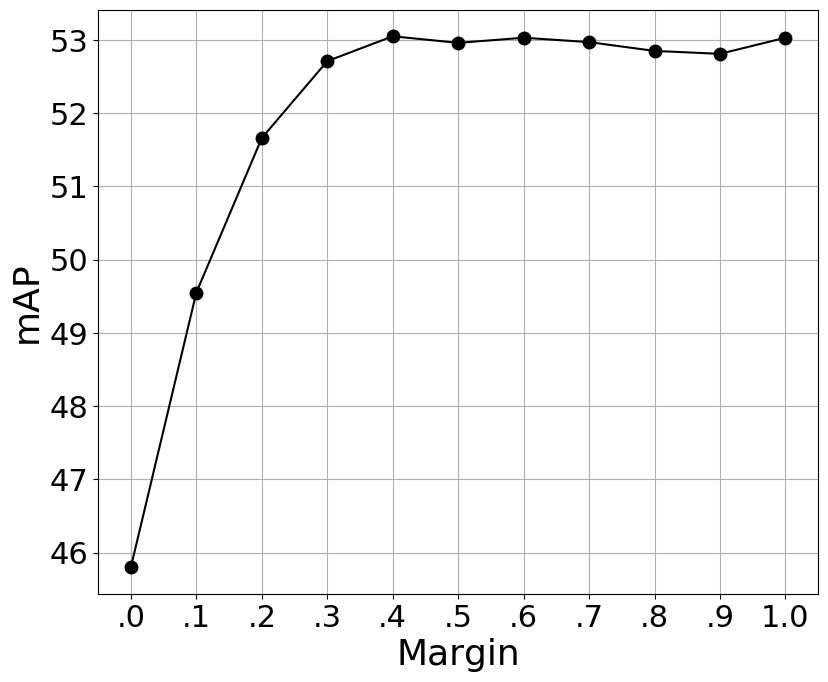}
  \caption{HPatches full split.}
\end{subfigure}
\begin{subfigure}[b]{0.49\linewidth}
\centering
  \includegraphics[width=0.98\linewidth]{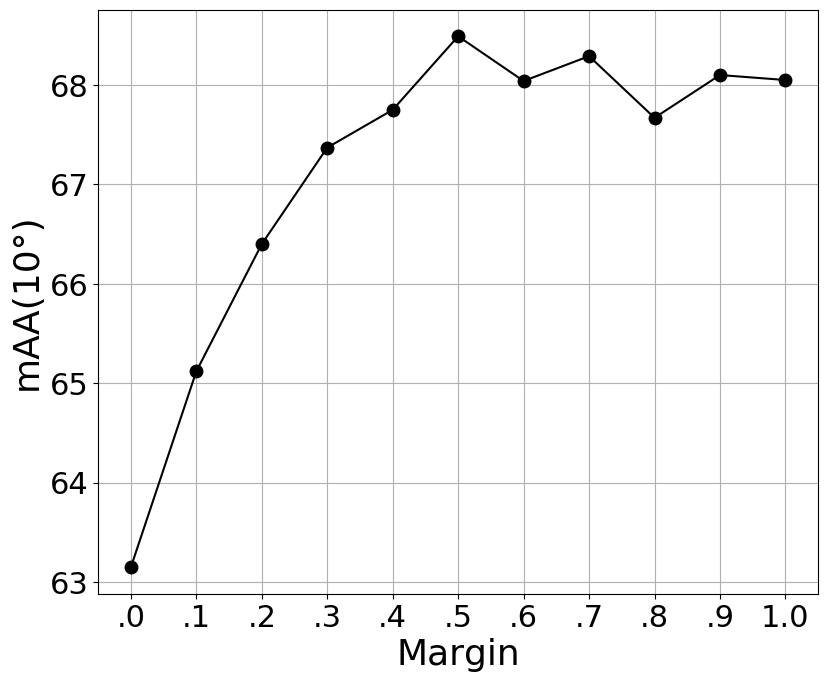}
  \caption{IMW PT, validation set.}
\end{subfigure}

\caption[HardNet performance depending on the margin applied during training.]{HardNet performance depending on the margin applied during training. Mean average precision (mAP) is measured on HPatches. In the IMW PT benchmark we measure mean average accuracy (mAA) with the threshold of 10\degree.}
% \caption{Matching score (mAP) for different margin used in the loss function.}
\label{fig:margin}
\end{figure}

\paragraph{Deciding on the Margin}
The purpose of this experiment is to determine the best value for the margin in the loss function. In Figure \ref{fig:margin} we can see that the performance roughly increases with the margin on both the CVPR and HPatches benchmarks. We can observe that the optimal value for margin is around 0.5 in case of the CVPR IMW benchmark. In HPatches the difference in evaluation between margins above 0.4 is almost negligible.

% Thanks to the observation, we suggest that the triplet margin hard-in-batch loss function can be simplified as 
% \begin{align*}
% L &= \frac{1}{n}\sum\limits_{i=1,n} d(p_{i,1},p_{i,2}) - min(d_1^i, d_2^i), \\
% d_1^i&=\min_{j=1,..n, j\neq i} d(p_{i,1}, p_{j,2}), \\
% d_2^i&=\min_{j=1,..n, j\neq i} d(p_{j,1}, p_{i,2}),
% \end{align*}

\paragraph{Sampling More Positives}
The hard-in-batch triplet margin loss can be generalized in the following way. Let $p_1,\dots,p_B$ be a set of patches in a batch with labels $l_1,\dots,l_B$. Corresponding patches have the same label. The generalized loss is then formulated as
\begin{align*}
\mathcal{L}&=\sum\limits_{l\in \textrm{unique(labels)}} \max(0,M+d_{\textrm{pos}}^l - d_{\textrm{neg}}^l), \\
d_{\textrm{pos}}^l&=\max\limits_{\substack{p_i, l_i=l \\ p_j, l_j=l \\ p_i \neq p_j}} d(p_i, p_j), d_{\textrm{neg}}^l=\min\limits_{\substack{p_i, l_i=l \\ p_j, l_j\neq l}} d(p_i, p_j).
\end{align*}
where $d_{\textrm{pos}}^l$ denotes the biggest distance within a set of positives and $d_{\textrm{neg}}^l$ denotes the smallest distance to the hardest negative outside of the set. Furthermore, we control the maximum number of positive patches in a batch, i.e. the batch size is variable and we only set its upper limit.

\begin{figure}[htb]
\centering

\begin{subfigure}[b]{0.49\linewidth}
\centering
  \includegraphics[width=0.99\linewidth]{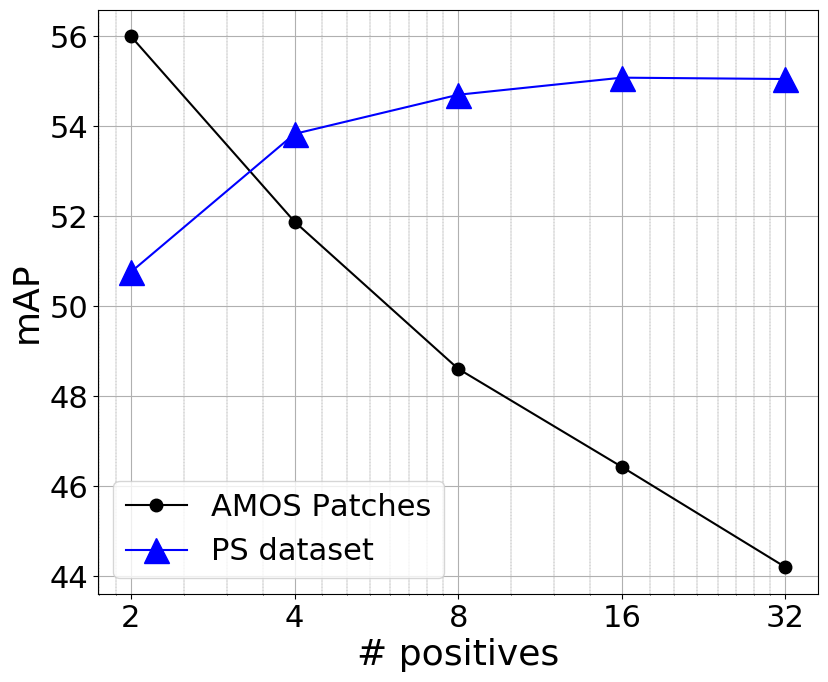}
  \caption{HPatches full split.}
\end{subfigure}
\begin{subfigure}[b]{0.49\linewidth}
\centering
  \includegraphics[width=0.99\linewidth]{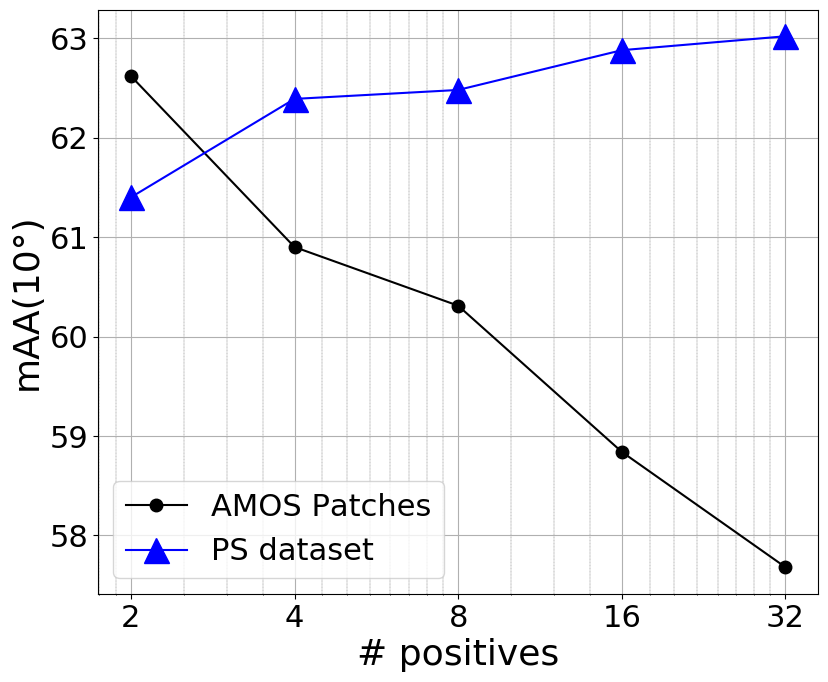}
  \caption{IMW PT, validation set.}
\end{subfigure}

\hfill\par

\caption[HardNet performance depending on the number of positives sampled in a batch during training.]{HardNet performance depending on the number of positives sampled in a batch during training. Different curves represent different training datasets. The measure is mean average precision (mAP) on HPatches and mean average accuracy (mAA) with the threshold of 10\degree on IMW PT.}
% \caption[Results achieved by HardNet as a function of the number of positives sampled in a batch during training.]{Results achieved by HardNet as a function of the number of positives sampled in a batch during training. Different curves represent different training datasets.}
\label{fig:positives}
\end{figure}

Evaluation is shown in Figure \ref{fig:positives} for two training datasets. Notice there is an opposite trend for the datasets - only for the PS dataset it is beneficial to sample more positives. We think the negative effect in AMOS Patches might be caused by an amount of noise in the dataset due to dynamic content such as moving cars, so that from a batch with more positives we actually sample false positives more often.

% \paragraph{ CDF-Based Dynamic Soft Margin Loss~\cite{Zhang_2019_ICCV}}
% Contrary to the triplet margin loss function, which uses a hard margin, the CDF-Based Dynamic Soft Margin loss uses a cumulative distribution function (CDF) to estimate the "hardness" of a triplet contributing to a batch. The CDF is computed by integrating all differences $d_{\textrm{pos}} - d_{\textrm{neg}}$ between the positive and negative pairs. The loss function is then formulated as $$\mathcal{L}=\frac{1}{B}\sum\limits_{i=1}^B \cdot (d_{\textrm{pos}}^i - d_{\textrm{neg}}^i) \cdot \textrm{CDF}(d_{\textrm{pos}}^i - d_{\textrm{neg}}^i),$$ i.e.  the fixed margin $M=1$ is replaced by a weight estimated from the batch. Such approach relieves the need to set the hard margin manually. As shown in table \ref{tab:softmargin}, the performance is inferior to our loss function, although according to the paper it should be the opposite. We ascribe this discrepancy to our different training setup (e.g. bigger batch size) and randomness in the training procedure. Training was performed on the Liberty dataset for 20 epochs, 5 million samples each.
% \input{Tabs/softmargin}

\section{Hyperparameters}
\label{sec:hyper}
We train our final models with learning rate~=~3 found by the LR Range test~\cite{smith2018disciplined}. Also, we use 1-Cycle style training~\cite{smith2018disciplined}. It comprises of two phases. First, the learning rate increases linearly to the initial value, then it decreases to 0 by following a cosine annealing schedule. In the following text we describe also other decisions we make regarding the training setup.

%for 20 epochs, 5 million samples each,

% In phase 1, the learning rates goes from lr_max/div_factor to lr_max linearly while the momentum goes from moms[0] to moms[1] linearly. In phase 2, the learning rates follows a cosine annealing from lr_max to 0

\paragraph{Augmentation}
During training, the patches are first horizontally and vertically flipped, both with probability 0.5. Then they are transformed to the input size. 

For AMOS Patches we apply random affine transformation and cropping to get patches of smaller size. First, random rotation from range $(\ang{-25},\ang{25})$, scaling from range $(0.8, 1.4)$ and shear are applied. Second, we crop a 64$\times$64 center of a patch. Then we crop a random area with scale in $(0.7,1.0)$ and aspect ratio in $(0.9,1.1)$ w.r.t. to the input patch and the patch is resized to 32$\times$32 pixels. These transformed patches are the input for training.

For Liberty we apply only resize transformation from 64$\times$64 to 32$\times$32 pixels. We tested other variants such as random affine transformation with or without Gaussian blur before downscaling, but none achieved better results.

\begin{figure}[htb]
\centering

\begin{subfigure}[b]{0.49\linewidth}
\centering
  \includegraphics[width=0.98\linewidth]{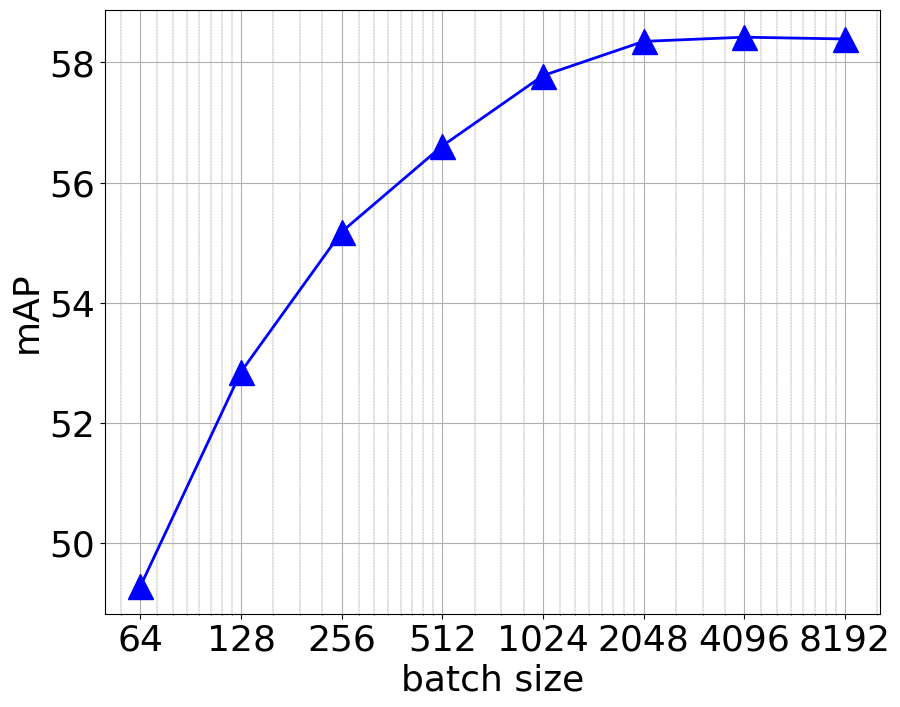}
  \caption{HPatches full split.}
\end{subfigure}
\begin{subfigure}[b]{0.49\linewidth}
\centering
  \includegraphics[width=0.98\linewidth]{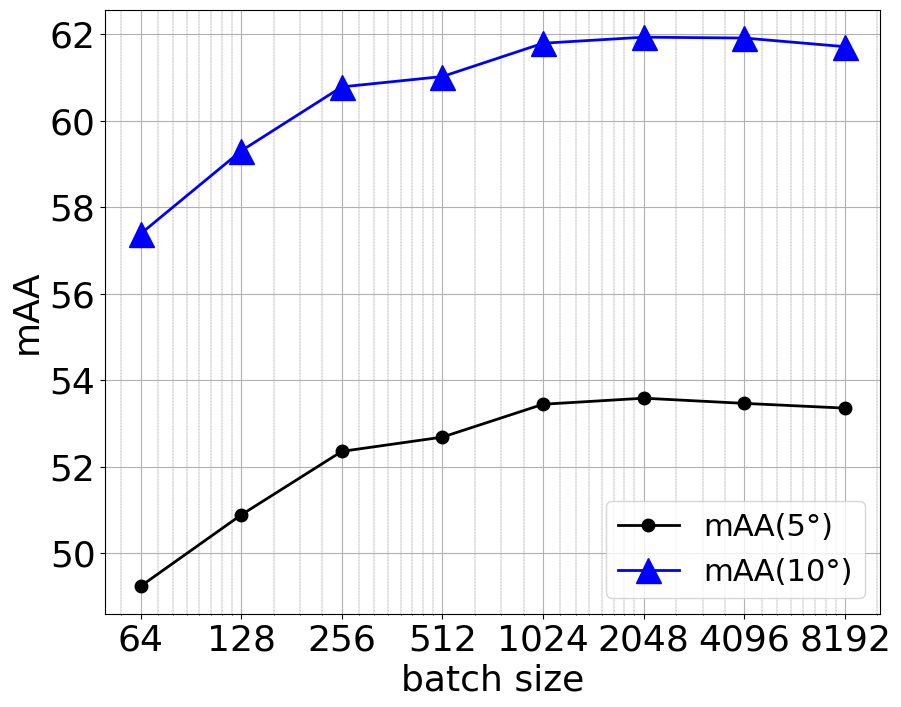}
  \caption{IMW PT, validation set.}
\end{subfigure}

\caption[HardNet performance depending on the training batch size.]{HardNet performance depending on the training batch size. Mean average precision (mAP) is measured on HPatches. In the IMW PT benchmark we measure mean average accuracy (mAA) with the threshold of 5\degree and 10\degree.}
\label{fig:bsize}
\end{figure}

\paragraph{Batch Size}
Batch size is an important factor in the setup of the training procedure. See in Figure \ref{fig:bsize} the influence of the batch size on the performance of the trained descriptor. We use HardNet architecture and train on Liberty for 10 epochs, 5 million samples each. Notice that there is an increase in performance, measured by mAP score on AMOS Patches and HPatches benchmarks, if we increase the batch size. It is likely that a higher batch size would be even more beneficial, but we are limited by the GPU memory. On a 32 GB machine the maximum batch size is approximately 8192. The best performance on the IMW PT benchmark is achieved for batch size around 4096. Interestingly, a further increase worsens the results on both HPatches and IMW PT.

\begin{figure}[htb]
\centering

\begin{subfigure}[b]{0.49\linewidth}
\centering
  \includegraphics[width=0.95\linewidth]{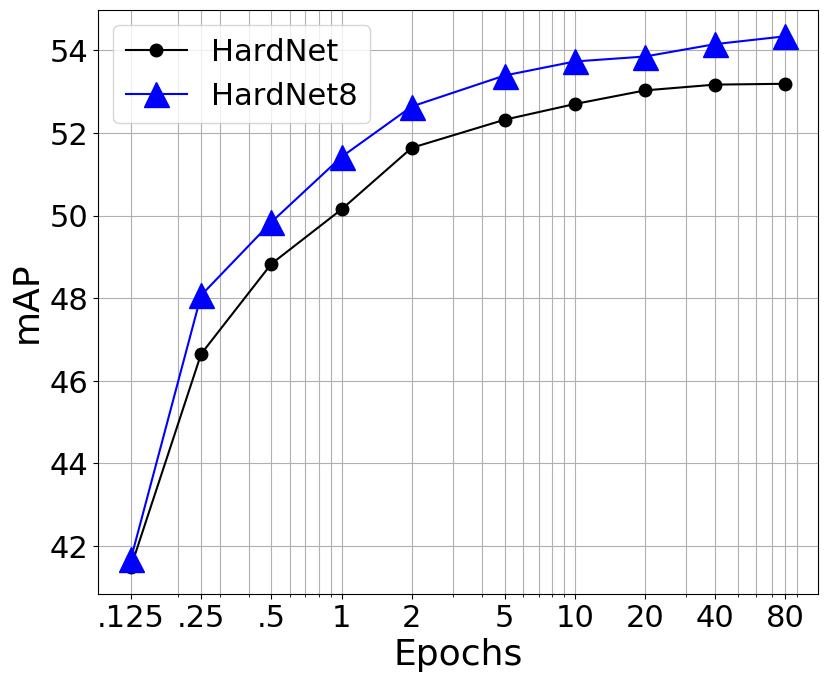}
  \caption{HPatches full split.}
\end{subfigure}
\begin{subfigure}[b]{0.49\linewidth}
\centering
  \includegraphics[width=0.98\linewidth]{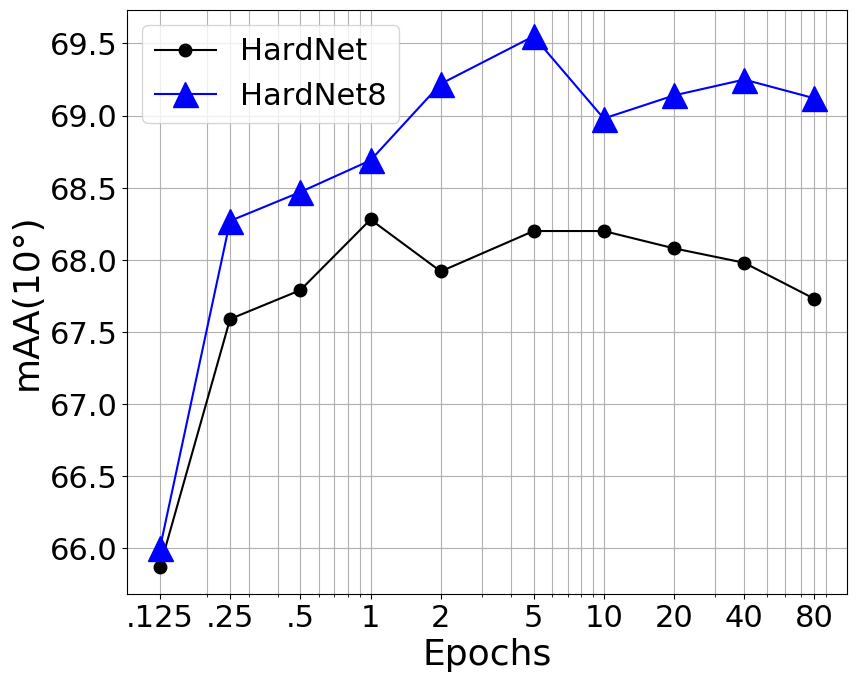}
  \caption{IMW PT, validation set.}
\end{subfigure}

\caption[Evaluation of HardNet trained for different number of epochs.]{Evaluation of HardNet trained for different number of epochs. Mean average precision (mAP) is measured on HPatches. In the IMW PT benchmark we measure mean average accuracy (mAA) with the threshold of 10\degree.}
% \caption[Mean average precision (mAP) and mean average accuracy (mAA) measures as functions of the number of epochs.]{Mean average precision (mAP) and mean average accuracy (mAA) measures as functions of the number of epochs. Note that all models were trained with random initialization. 1 epoch $\approx$ 5 million samples.}
\label{fig:epochs}
\end{figure}

\paragraph{Number of Epochs}
To decide on the number of epochs for which we train HardNet and HardNet8 models on the Liberty dataset for a different number of epochs and observe its performance. See in Figure \ref{fig:epochs} that it steadily increases on HPatches with the number of epochs, while it peaks at 1 epoch in the IMW PT benchmark and then decreases.

\chapter{Evaluation on Standard Benchmarks}
We evaluate two HardNet8 variants which use the improvements validated in the previous chapter. The first, HardNet8-Univ, was trained on Liberty and AMOS Patches datasets for 20 epochs, 5 million samples each, with the generalized loss function with sampling of 2 positive patches, batch size~=~3072.
% we conjecture
% finetune with your dataset - advice to user..., start with
% By combining all the improvements we found, we have created two final models. The first, HardNet8-Univ, uses the HardNet8 architecture and was trained on Liberty and AMOS Patches datasets for 20 epochs, 5 million samples each, with the generalized loss with sampling of 2 positive patches, batch size~=~3072.

The second model, HardNet8-PT, has 512 output channels and was trained on Liberty, Notredame and Colosseum Exterior datasets for 40 epochs, 5 million samples each, batch size~=~9000. The embeddings from the model are compressed by PCA -- which is fitted on Liberty -- to dimensionality of 128.

Evaluation is performed on the test set of IMW PT -- the previous experiments were evaluated on the validation set. HPatches and AMOS Patches comprise of a single test split which is used both in the experiments and final evaluation.

Results are shown in Table \ref{tab:results}. HardNet8-Univ  works reasonably well on a wide range of conditions, be it viewpoint or illumination change, and improves state-of-the-art on the standard benchmarks. It is only outperformed by GeoDesc in the viewpoint split of HPatches. HardNet8-PT further improves the performance on the IMW PT benchmark in the Stereo 8k task.
\begin{table}[htb]
\noindent %%%%%%%%%%%%
\centering
\caption[Final evaluation of the proposed descriptors.]{Final evaluation of the proposed HardNet8 descriptors. Mean average precision (mAP) is measured on HPatches and AMOS Patches. On the IMW PT benchmark we measure mean average accuracy (mAA) with the threshold of 10\degree.}
% \caption{Final evaluation on HPatches and AMOS Patches. The measure is mean average precision (mAP).}
\makebox[\textwidth]{ %%%%%%%%%%%%
\footnotesize
\begin{tabular}{lcccccc}
\toprule
Model & Trained on &\multicolumn{3}{c}{HPatches subset} & AMOS &IMW PT  \\
& &illum & view & full &Patches& Stereo 8k  \\
\midrule
SIFT &-- & 23.33 & 28.88 & 26.15 & 37.08 & 45.84\\
HardNet & Liberty & 49.86&	55.63&	52.79&	43.32&55.43 \\
SOSNet & Liberty &50.66&	56.73&	53.75 & 43.26&55.87 \\
HardNetPS & PS Dataset & 48.55&	67.43&	58.16 & 31.83 & 50.51 \\
GeoDesc & GL3d~\cite{shen2018mirror} & 50.52 & \textbf{67.48} & 59.15 & 25.50&51.11 \\
\midrule
HardNet8-Univ&AMOS+Liberty &\textbf{58.20}&63.60&\textbf{60.95}&\textbf{47.20}&56.22 \\
HardNet8-PT& Lib+Coloss+Notre & 51.04&55.81&53.46 & 45.01&\bf{57.58}  \\
\bottomrule
\end{tabular}
} %%%%%%
\label{tab:results}
\end{table}
% CV-SIFT & .2875 & .4584 \\
% CV-$\sqrt{\text{SIFT}}$ & .3149 & .4930 \\
% \midrule
% D2-Net (SS) & .1355& .2228  \\
% D2-Net (MS) & .1813 & .2506\\
% DoG-GeoDesc & .3662 & .5111  \\
% ContextDesc & .3510& .5098  \\
% DoG-SOSNet & .3976 & .5587  \\
% DoG-LogPolarDesc & 4115& .5340\\
% DoG-HardNet & \bf{.4029}& .5543 \\
% \midrule
% DoG-HardNet8-PT (ours)  &.3865    &.\bf{5758} \\
% DoG-HardNet8-Univ (ours) &.3704    & .5622 \\
%%%%%%%%%%%%%%%%%%%%%%%%%%%%%%55
%\begin{table}[htb]
%% \noindent %%%%%%%%%%%%
% \centering
% \caption{Final evaluation on HPatches and AMOS Patches. The measure is mean average precision (mAP).}
% \makebox[\textwidth]{ %%%%%%%%%%%%
% \begin{tabular}{lccccc}
% \toprule
% Model & Trained on &\multicolumn{3}{c}{HPatches subset} & AMOS Patches  \\
% & &illum & view & full  \\
% \midrule
% HardNet8-PT (ours) & Lib+Colosseum+Notre & 51.04&	55.81&	53.46 \\
% HardNet8-Univ (ours) & AMOS+Liberty &\textbf{58.20}&63.60&\textbf{60.95}&\textbf{47.20} \\
% \midrule
% SIFT &-- & 23.33 & 28.88 & 26.15 & 37.08 \\
% HardNet & Liberty & 49.86&	55.63&	52.79&	43.32 \\
% SOSNet & Liberty &50.66&	56.73&	53.75 & 43.26 \\
% HardNetPS & PS Dataset & 48.55&	67.43&	58.16 & 31.83 \\
% GeoDesc & GL3d~\cite{shen2018mirror} & 50.52 & \textbf{67.48} & 59.15 & 25.50 \\
% \bottomrule
%
% \end{tabular}
% } %%%%%%
% \label{tab:results}
% \end{table}
%

Apart from benchmarking, we also conjecture that significantly better results can be achieved for a real-world task if HardNet8 is retrained on a specific combination of datasets, suitable for the use case -- e.g. AMOS Patches for illumination changes with no viewpoint change, while Liberty, Notredame, Colosseum Exterior and others for purely viewpoint changes.
%
% This dataset\footnote{available at https://github.com/vcg-uvic/image-matching-benchmark} was recently created for the upcoming CVPR Image Matching Workshop and is a part of extensive evaluation software. Currently it contains 13 scenes for training, 3 for validation and 10 for testing. COLMAP \cite{schonberger2016structure} SfM software was used to create 3D reconstructions with verified patch correspondences. Each folder with a 3D reconstructed scene contains 2D to 3D point correspondences. It does not provide information about scale and rotation and the extracted patches cannot be easily normalized. However, we have observed that it is not a big issue in terms of the performance of the trained descriptor. Because the images are obtained from Phototourism data, it covers some changes in weather and lighting. The number of extracted patches is proportional to the total number of keypoints in all images and varies highly across the scenes. E.g. Brandenburg Gate contains over 2 million of patches, while Trevi Fountain contains over 15 million.
% \chapter{Image-level transformation}
\chapter{Conclusion}
\label{sec:conclusion}
We have studied various factors, which influence the performance of a local feature descriptor, and made improvements to the HardNet descriptor.

AMOS Patches dataset is presented, which improves robustness of trained local feature descriptors to illumination and appearance changes. It is based on registered images from selected cameras from the AMOS dataset. Based on our experience we give recommendations related to using webcams for the local descriptor learning: (i) When picking cameras for training manually, a small and diverse subset is better than a bigger one with similar views or imprecise alignment of images; (ii) When sampling patches for a batch during training, using less source cameras is beneficial.

% improvement by x%
% It has both the training and testing split. 
%We provide the AMOS Patches dataset for robustification of local feature descriptors to illumination and appearance changes. 

% \begin{itemize}
%     % \item Scene parsing methods do not work well in outdoor webcams. The precision of the near state-of-the-art network \cite{zhou2018semantic} is not satisfactory.
%     % \item For camera selection we recommend to adopt strict "quality" criteria and be prepared to loose many suitable cameras in the process.
%     \item When picking cameras for training manually, a small and diverse subset is better than a bigger one with similar views or imprecise alignment of images.
%     % \item Better results and smaller dataset are achieved by sampling images for extraction of patches in a determnistic way, e.g. using the Hessian detector, rather than using some random distribution.
%     \item When sampling patches for a batch during training, using less source cameras is beneficial.
    
%     %use as few source cameras as possible.
% \end{itemize}

We present a study on combining and filtering heterogeneous datasets into a single compact dataset for the learning of a local descriptor. A technique for dataset reduction was described, we showed that training on a reduced dataset of by an order of magnitude smaller size yields comparable performance. Enriching data modality by switching to RGB or concatenating patches with estimated mono-depth maps leads to limited or no improvement.

%  and provide relevant recommendations

 HardNet8 architecture is presented, achieving consistently better results than the original HardNet. Other architectural and training choices were explored. We show ResNet-like architectures and models found by the DARTS algorithm are inferior to those of the VGG-like models. Similar observation was made about switching to local pooling or increasing the receptive field. Conversely, bigger training batch size is beneficial. Also, using PCA to compress the network outputs leads to better results on the IMW PT benchmark.
 
%  better results on HPatches, AMOS Patches and IMW PT benchmarks
 
%  Bigger batch size during training has positive influence on the performance of the descriptor

We also present a set of the findings, which might be useful for the practicioners,  such as that the hardest negative often goes from the same image, training on more datasets might deteriorate the descriptor quality and others.

By combining all the improvements we found,  we have trained two models\footnote{Trained models available at https://github.com/pultarmi/HardNet\_MultiDataset}. One improves state-of-the-art on the HPatches, AMOS Patches and IMW PT. The other, specifically tuned to the IMW PT, further improves results on that dataset at the cost of being slightly worse on HPatches and AMOS patches.

\medskip

\printindex

\bibliographystyle{abbrv}
\bibliography{ctutest}

\end{document}